\documentclass[letterpaper]{article} %
\usepackage{aaai23}  %
\usepackage{times}  %
\usepackage{helvet}  %
\usepackage{courier}  %
\usepackage[hyphens]{url}  %
\usepackage{graphicx} %
\usepackage{natbib}  %
\usepackage{caption} %
\usepackage{algorithm}
\usepackage{algorithmic}
\usepackage[T1]{fontenc}

\usepackage[colorlinks]{hyperref}       %
\usepackage{url}            %
\usepackage{booktabs}       %
\usepackage{amsfonts, amssymb}       %
\usepackage{booktabs} %
\usepackage{amsmath}
\usepackage{nicefrac}       %
\usepackage{microtype}      %
\usepackage{xcolor}  
\usepackage{graphicx}
\usepackage{float}
\usepackage{subcaption}
\usepackage{diagbox}
\usepackage{multirow}

\newcommand*\samethanks[1][\value{footnote}]{\footnotemark[#1]}

\usepackage{newfloat}
\usepackage{listings}
\DeclareCaptionStyle{ruled}{labelfont=normalfont,labelsep=colon,strut=off} %
\floatstyle{ruled}
\newfloat{listing}{tb}{lst}{}
\floatname{listing}{Listing}
\title{ACE: Cooperative Multi-agent Q-learning with 
Bidirectional Action-Dependency}
\author {
    Chuming Li\textsuperscript{\rm 1, 2}\thanks{Equal contribution},
    Jie Liu\textsuperscript{\rm 2}\samethanks, 
    Yinmin Zhang\textsuperscript{\rm 1, 2}\samethanks,
    Yuhong Wei\textsuperscript{\rm 3}, 
    Yazhe Niu\textsuperscript{\rm 2, 3}, 
    Yaodong Yang\textsuperscript{\rm 4}\thanks{Corresponding author}, \\
    Yu Liu\textsuperscript{\rm 2, 3}\samethanks, 
    Wanli Ouyang\textsuperscript{\rm 1, 2}
}
\affiliations {
    \textsuperscript{\rm 1} The University of Sydney, SenseTime Computer Vision Group, Australia,
    \textsuperscript{\rm 2} Shanghai Artificial Intelligence Laboratory, \\
    \textsuperscript{\rm 3} SenseTime Group LTD,
    \textsuperscript{\rm 4} Institute for AI, Peking University \\
    chli3951@uni.sydney.edu.au, liujie@pjlab.org.cn, \{yinmin.zhang, wanli.ouyang\}@sydney.edu.au,
    \{weiyuhong, niuyazhe\}@sensetime.com,
    yaodong.yang@pku.edu.cn, 
    liuyuisanai@gmail.com,
}

\usepackage{bibentry}

\begin{document}

\maketitle

\begin{abstract}

Multi-agent reinforcement learning (MARL) suffers from the non-stationarity problem, which is the ever-changing targets at every iteration when multiple agents update their policies at the same time. Starting from first principle, in this paper, we manage to solve the non-stationarity problem by proposing bidirectional action-dependent Q-learning (ACE). 
Central to the development of ACE is the sequential decision making process wherein only one agent is allowed to take action at one time.
Within this process, each agent maximizes its value function given the actions taken by the preceding agents at the inference stage.  
In the learning phase, each agent minimizes the TD error that is dependent on how the subsequent agents have reacted to their chosen action. 
Given the design of bidirectional dependency, ACE effectively turns a multi-agent MDP into a single-agent MDP.
We implement the ACE framework by identifying the proper network representation to formulate the action dependency, so that the sequential decision process is computed implicitly in one forward pass. 
To validate ACE, we compare it with strong baselines on two  MARL benchmarks. 
Empirical experiments demonstrate that ACE outperforms the state-of-the-art algorithms on Google Research Football and StarCraft Multi-Agent Challenge by a large margin. 
In particular, on SMAC tasks, ACE achieves 100\% success rate on almost all the hard and super hard maps.
We further study extensive research problems regarding ACE, including extension, generalization and practicability. \href{https://github.com/opendilab/ACE}{Code} is made available to facilitate further research.

\end{abstract}

\section{Introduction}
\vspace{-0.5ex}
Cooperative multi-agent reinforcement learning (MARL) aims to learn a good policy that controls multiple agents and maximizes the cumulative return in a given task. It has great potential in various real-world tasks, such as robot swarm control~\citeyear{swarm}, autonomous driving~\citeyear{smarts,safe_ad} and multi-player games~\citeyear{mobo,pettingzoo}. A major challenge of MARL is the complex joint action space. In multi-agent tasks, the joint action space increases exponentially with the number of agents. Hence, for the sake of scalability, existing MARL algorithms usually learn an individual policy to select the action for every single agent. In MARL algorithms, the reward signal is affected by other agents' behavior. However, the environment of multi-agent task is non-stationary~\citeyear{non_stationarity,multireview} to every single agent, where the policies of agents keep changing during the learning process. This non-stationary problem breaks the Markov assumption in single-agent RL algorithms and causes endless adaptation of multiple agents according to each other's change of policy.
In value-based methods, the non-stationary problem shows up as that the value of the individual action can not be estimated accurately.

To solve the non-stationary problem, we introduce bidirectional action-dependency to estimate the action value of every single agent accurately. 
We cast multi-agent decision-making process as a sequential decision-making process, where only one agent makes a decision at a time.
In this sequential process, the bidirectional action-dependency is embodied in two aspects. 
In the forward direction, the evaluation of an agent's action value is dependent on the preceding agents' actions in the decision-making sequence.
While in the backward direction, the target to update an agent's action value is dependent on how subsequent agents react to the preceding actions. 
We formulate this bidirectional dependence by transforming a multi-agent Markov Decision Process~(MMDP)~\citeyear{markov} into a single-agent Markov Decision Process~(MDP), called sequentially expanded MDP~(SE-MDP). In SE-MDP, a decision $\boldsymbol{a}^t$ based on a state $s^t$ is expanded to multiple intermediate states $\left[s^t_{a_1},...,s^t_{a_{1:n}}\right]$, named SE-state. The SE-state $s^t_{a_{1:i}}$ is defined as the state $s^t$ in the original MMDP along with the decisions $a_{1:i}$ made by the preceding agents. Only one agent makes a decision at each SE-state. After each agent makes the decision, the state transits to the next one, which includes the new decision. This transformation validates that the proposed bidirectional action-dependency does circumvent the non-stationary problem.

\begin{figure*}[t]
    \centering
    \includegraphics[width=0.8\textwidth]{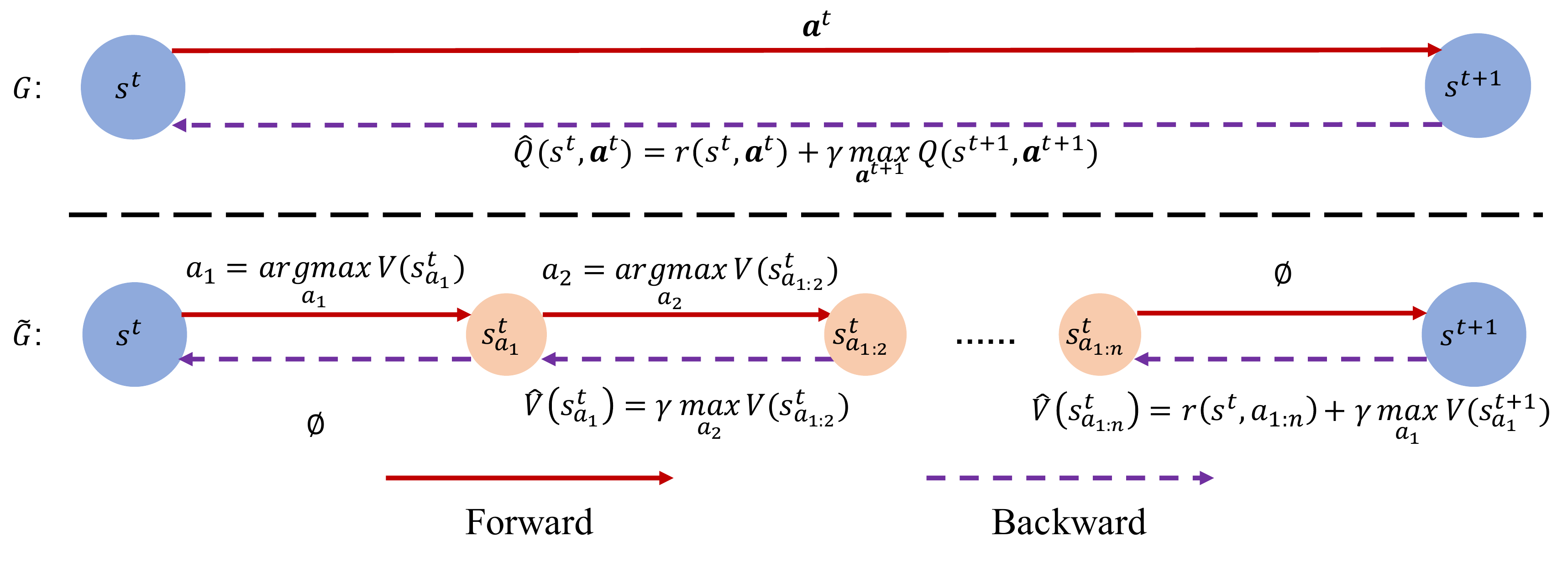}
\vspace{-3ex}
    \caption{Comparison between the original MMDP (above) and the transformed SE-MDP (below). A single transition in MMDP is expanded to $n$ sequentially expanded states in SE-MDP.}
\vspace{-3ex}
    \label{fig:mdp}
\end{figure*}

With the introduced bidirectional dependency, we propose a simple but powerful method, called \textit{bidirectional ACtion-dEpendent deep Q-learning}~(ACE). ACE is compatible with the abundant Q-learning methods for single-agent tasks, and naturally inherits their theoretical guarantee of convergence and performance. For practical implementation, we identify an efficient and effective network representation of the SE-state. 
We first generate the embeddings for all units in the task as well as the embeddings for their available actions. 
Then, we combine the embedding of each unit with the corresponding action embedding to construct the embedding for every SE-state. 
This design is quite efficient, because the embeddings of all SE-states along a sequential decision-making process are constructed with additive combination among the same set of unit and action embeddings. This set is computed only once before every sequential decision-making process, and the additive combination brings in negligible cost. 
Moreover, an interaction-aware action embedding is developed to describe the interaction among units in the multi-agent task, which further improves the performance of ACE.

We evaluate the performance of ACE on both a toy case and complex cooperative tasks. In the toy case, ACE demonstrates its advantage in converging to the optimal policy against the popular value-factorization methods. Because it bridges the gap of the optimal actions between the joint and individual Q-function, which widely exists in value-factorization methods. For complex tasks, we choose two benchmark scenarios in Google Research Football (GRF)~\citeyear{grf} environment and eight micromanagement tasks in StarCraft Multi-Agent Challenge (SMAC)~\citeyear{smac}. Empirical results show that ACE significantly outperforms the state-of-the-art algorithms on GRF, and achieves higher sample efficiency by up to 500\%. On SMAC, ACE achieves 100\% win rates in almost all the hard and super-hard maps. Other advantages of ACE are verified with comprehensive experiments, including generalization, extension and practicability. Surprisingly, ACE also indicates better generalization performance compared with other baselines when transferred to a new map with a different number of agents in SMAC.

\section{Related Work}
\vspace{-0.5ex}
To solve the widespread cooperation tasks, many multi-agent reinforcement learning~(MARL) algorithms have been proposed recently. According to the extent of centralization, these works can be divided into two categories, independent learning scheme and action-dependent learning scheme.

First, many works tend towards a fully independent learning scheme~\citeyear{multisurvey}, where agents make decisions with their independent value functions or policies. One typical category assigns independent actor to each agent by directly transferring the actor-critic methods to multi-agent scenarios~\citeyear{coma, maddpg, mappo, noisy_mappo}.
Another line is value-based methods~\citeyear{iql, vdn, qmix, wqmix, qtran, qtran++, collaq, wang2020qplex}.
To avoid the non-stationary problem, they usually develop different factorized value functions following the IGM principle~\citeyear{qtran}, which requires that the individually optimal actions are consistent with the jointly optimal actions. We remark that existing value factorization methods following the IGM principle either suffer from the structural constraints, like VDN and QMIX, or introduce secondary components along with additional hyperparameters, like QTRAN, WQMIX and QPLEX. 
However, the optimal joint action often changes due to the discovery of a better policy, resulting in the mismatch between the optimal joint Q function and individual functions during training. This means that individual Q functions require more iterations to recover the satisfaction of IGM, and the policy explores the environment with sub-optimal actions, leading to low sample efficiency.
To avoid the issues, this paper focuses on directly estimating the value of each action, rather than following the IGM principle to construct factorization function classes.

\begin{figure*}[t]
    \centering
    \includegraphics[width=0.75\textwidth]{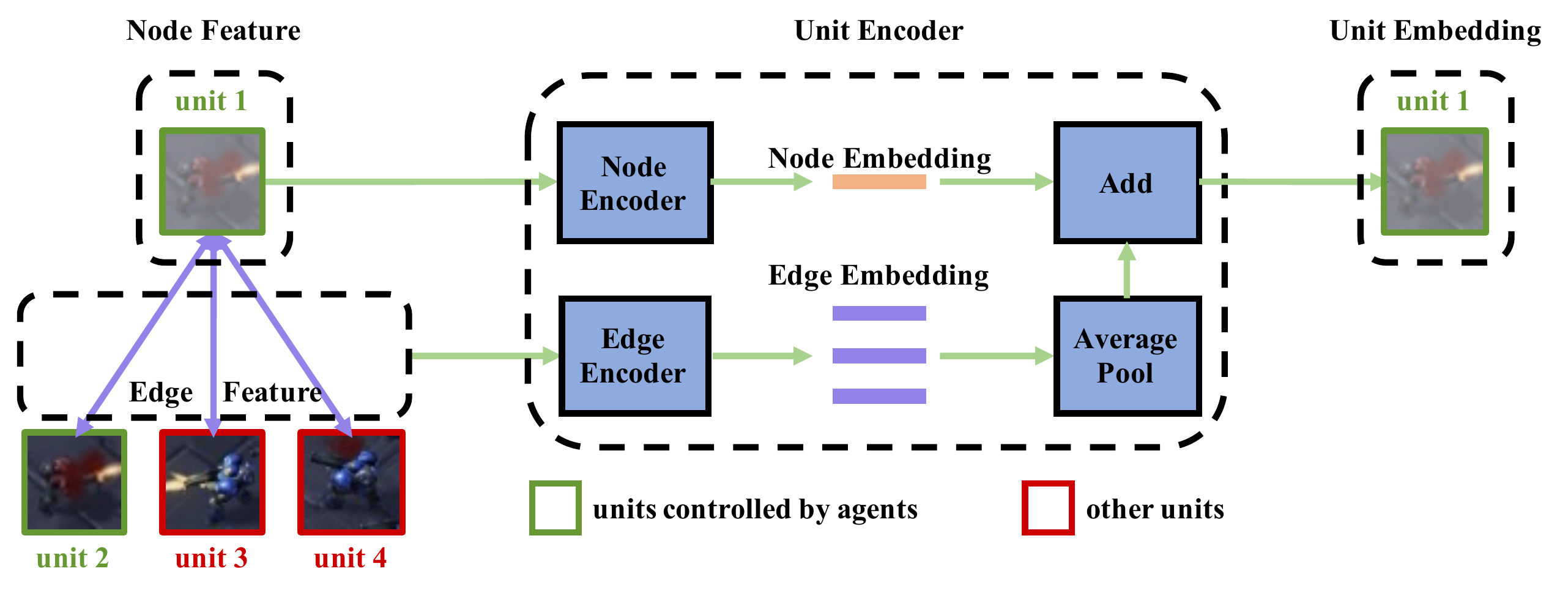}
\vspace{-3ex}
    \caption{The schematic of the unit encoder. The node embedding is obtained by the node encoder, and the edge embedding (for the unit and its interacted units) is obtained from the edge encoder. The average-pooled edge embedding is added to the node embedding to provide unit embedding.}
\vspace{-4ex}
    \label{fig:unit_encoder}
\end{figure*}

Second, the action-dependent learning scheme~\citeyear{multiagent_rollout,happo,gcs,kuba2021settling, ye2022towards, kuba2022heterogeneous, fu2022revisiting} is more centralized. One perspective is action-dependent execution, where the agent makes decisions with dependency on other agents’ actions. CGS~\citeyear{gcs} proposes a graph generator to output a directed acyclic graph which describes the action dependency. Each node in the graph represents an agent whose policy is dependent on the action of agents on its parent nodes. However, each agent's decision is only dependent on part of the previous agents in the topological sort of the generated DAG, and the policy update is independent on the reaction of the subsequent agents. It means the non-stationary effect is not totally removed. In another perspective, action-dependency is introduced in policy update rather than execution. Multi-agent rollout algorithm~\citeyear{multiagent_rollout} and HAPPO~\citeyear{happo} follow a update to sequentially update the policy of each agent with the others fixed, thus avoiding the conflicting update directions of individual policy updates. This paradigm is an implicit rather than full action-dependency, because the policy does not explicitly depends on the actions of the preceding agents.
As an extra difference, ACE is the first value-based MARL method that achieves remarkable performance following the action-dependent learning scheme. More elaborate discussion with relative works are referred to the Appendix.

\vspace{-1ex}
\section{Problem Formulation}
\vspace{-0.5ex}
In this paper we take Multi-agent Markov Decision Process (MMDP)~\citeyear{markov} to model cooperative multi-agent tasks. An MMDP is a tuple 
$\mathcal{G}=\left\langle \mathcal{S}, \mathcal{N}, \mathcal{A}, P, r, \gamma \right\rangle$, where $\mathcal{S}$ is the space of global state and $\mathcal{N}$ is the set of $n$ agents. $\mathcal{A}\equiv \mathcal{A}_1 \times, ..., \times \mathcal{A}_n $ is the joint action space consisting of each agent's action space $\mathcal{A}_i$. At each step, the global state $s$ is transformed to each agent $i$'s input, and each agent $i$ selects an action $a_i \in \mathcal{A}_i$. Then, with the joint action $\boldsymbol{a}=\left[a_1,...,a_n\right]$ and the transition function $P\left(s^{\prime}|s,\boldsymbol{a}\right)$, the process transits to the next state $s^{\prime}$ and returns a reward $r\left(s,\boldsymbol{a}\right)$. The target we consider is to learn an optimal policy $\boldsymbol{\pi}\left(\boldsymbol{a} \mid s\right)$ which maximizes the expected return $\mathcal{R} = \mathbb{E}_{\boldsymbol{\pi}}\left[\sum_{t=0}^{\infty} \gamma^t r\left(s^t,\boldsymbol{a^t}\right)\right]$.

\section{Method}
\subsection{Bidirectional Action-Dependency}
\vspace{-0.5ex}
In this section, we consider a sequential decision-making scheme: all agents make decisions sequentially. The bidirectional action-dependency has two directions. In the forward direction, each agent's decision depends on the state and their preceding agents' actions. Inversely, in the backward direction, the update of the Q-value for an agent's action depends on how its successor reacts to the preceding actions. 

We formalize this bidirectional dependency by transforming the original MMDP $\mathcal{G}$ into a single agent MDP $\widetilde{\mathcal{G}}$. In $\widetilde{\mathcal{G}}$, the state transits along the decision-making sequence. Specifically, a intermediate transition happens each time when a single agent in the sequence selects its action. The intermediate state is defined as the original state $s^t$ along with the actions of the agents which have made their decisions, denoted as $s_{a_{1:i}}^t$. At each intermediate transition, an agent $i$ receives its intermediate state $s_{a_{1:i-1}}^t$ and produces its action $a_i$, then the intermediate state intermediately transits to $s_{a_{1:i}}^t$ with a reward $0$. After the last agent $n$ makes decision and the intermediate state intermediately transits to $s_{a_{1:n}}^t$, a psuedo agent produces an empty action $\empty$ and the intermediate state transits from $s_{a_{1:n}}^t$ to $s^{t+1}$, with the reward $r\left(s^t,\boldsymbol{a}^t\right)$ defined in the original MMDP $\mathcal{G}$. With the above definition, a transition $\left(s^t,\boldsymbol{a}^t,r\left(s^t,\boldsymbol{a}^t\right),s^{t+1}\right)$ of $\mathcal{G}$ is expanded into a sequence of intermediate transitions $\left(s^t,a_1^t,0,s_{a_1}^t\right),\left(s_{a_1}^t,a_2^t,0,s_{a_{1:2}}^t\right),...,(s_{a_{1:n-1}}^t,a_n,r (s^t, \boldsymbol{a}^t),$  $ s^{t+1})$ in $\widetilde{\mathcal{G}}$. We define $\widetilde{\mathcal{G}}$ as the sequential expansion of $\mathcal{G}$ and name this MDP as sequentially expanded MMDP (SE-MMDP). Similarly, we define the intermediate state $s_{a_{1:i}}$ as sequentially expanded state (SE-state), of which the space is represented by $\widetilde{\mathcal{S}}$.

As depicted in Figure~\ref{fig:mdp}, the formulation of SE-MDP validates that the bidirectional action-dependency does circumvent the non-stationary problem. In SE-MDP, the forward dependency is manifested in that the preceding actions are incorporated in the SE-state. It means the changeable behavior of the preceding agents are tracked in the value estimation of each SE-state. As for the backward dependency described by dashed lines in Figure~\ref{fig:mdp}, the target value of an agent's action $a_i$ in the Bellman operator depends on its successor's reaction to the preceding actions, \textit{i.e.,} the best selection of $a_{i+1}$, which also tracks the successor's behavior.

\subsection{Bidirectional Action-Dependent Q-learning}
\vspace{-0.5ex}
The formulation of sequential expansion $\widetilde{\mathcal{G}}$ circumvents the non-stationary problem, which enables us to easily adopt different single-agent algorithms to solve $\widetilde{\mathcal{G}}$. Based on the formulation of sequential expansion, this section introduces the proposed bidirectional ACtion-dEpendent Q-learning~(ACE), which transfers existing single-agent value-based methods to multi-agent scenarios with minimalist adaptation and inherits their theoretical guarantee of convergence and performance.

\begin{figure*}[t]
    \centering
    \includegraphics[width=0.95\textwidth]{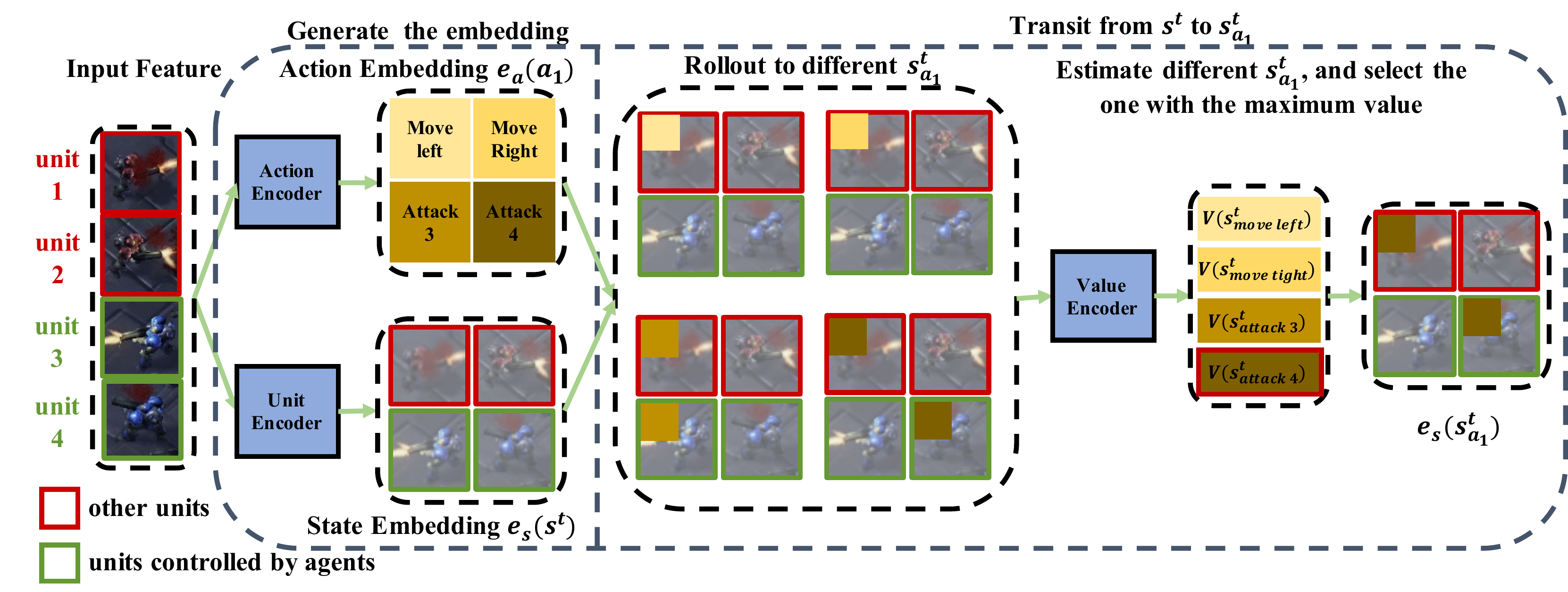}
\vspace{-2ex}
    \caption{Schematic of the pipeline of ACE, which takes SMAC as an instance. There are four units in the map. Units 1 and 2 are controlled by the RL agent, and units 3 and 4 are enemies controlled by the environment. At first, the initial state embedding is generated, consisting of the initial embedding for all units obtained from the unit encoder, as well as the action embedding of all actions obtained from the action encoder (only the action embedding of unit 1 is shown in the figure, where actions \emph{attack 3} and \emph{attack 4} mean unit 1 attacking unit 3 and 4 respectively). Then, agent (unit) 1 is the first one to make the decision, thus its action embeddings are incorporated into the initial unit embeddings to rollout to the embeddings of different new SE-states $e\left(s_{a_1}^t\right)$ (4 rolled out SE-states in the figure). Afterwards, all of these new SE-states are evaluated by the value encoder. Finally, the SE-state with the maximum value is retained and used by the next rollout for the action of agent 2.}
\vspace{-4ex}
    \label{fig:transition}
\end{figure*}

Value-based methods usually learn the function $Q: \mathcal{S} \rightarrow \mathbb{R}^{\left| A \right|}$ to build the mapping from the state to the estimated return of actions, and select the action with the maximum $Q$ value during execution. However, in SE-MDP, once making the decision $a_{i+1}$ for the $i+1$th agent on the current SE-state $s_{a_{1:i}}^t$, we can direct intermediate transition to the next SE-state $s_{a_{1:i+1}}^t$ without interacting with the environment. Hence, we take a step forward and use the value function $V: \widetilde{\mathcal{S}} \rightarrow \mathbb{R}$ to estimate the return of the SE-state rather than the action, and use the values $V\left(s_{a_{1:i+1}}^t\right)$ of all possible next SE-states rolled out via different actions $a_{i+1}$ to select the optimal action.

\textbf{Decision with Rollout} Specifically, to make decision at an SE-state $s_{a_{1:i}}^t$, we use agent $i+1$'s action space $A_{i+1}$ to roll out to all possible next SE-states $s_{a_{1:{i+1}}}^t$, and select the action $a_{i+1}^t = \mathop{\arg\max}_{a_{i+1}} \ V\left(s_{a_{1:i}, a_{i+1}}^t\right)$, which leads to the next SE-state with the optimal value $V\left(s_{a_{1:i+1}}^t\right)$.

\textbf{Update with Rollout} Our value function $V$ is updated by the standard Bellman backup operator in single agent RL. At an SE-state $s_{a_{1:i}}^t$, to obtain the target value to update the value $V\left(s_{a_{1:i}}^t\right)$, we also rollout to all possible next SE-states $s_{a_{1:{i+1}}}^t$, estimate their values $V\left(s_{a_{1:i+1}}^t\right)$ and select the maximum value as the target value. For the final SE-state $s_{a_{1:{n}}}^t$ in a decision sequence, we roll out at the first SE-state $s^{t+1}$ in the next decision sequence, i.e., the next state in the original MMDP $\mathcal{G}$. The update of $V$ is formalized as Eq~\ref{bellman_update}, with $\hat{V}\left(s_{a_{1:i}}^t\right)$ denoting the Bellman target of $V\left(s_{a_{1:i}}^t\right)$.
\begin{gather}
\label{bellman_update}
    \hat{V}\left(s_{a_{1:i}}^t\right)=
    \begin{cases}
        \mathop{\max}_{a_{i+1}} \ \gamma V\left(s_{a_{1:i}, a_{i+1}}^t\right),  & \text{if $i < n$} \\
        \mathop{\max}_{a_{1}} \ r\left(s^t,a_{1:n}\right)
         + \gamma V\left(s_{a_{1}}^{t+1}\right), & \text{if $i = n$}
    \end{cases}
\end{gather}
\vspace{-4mm}
\vspace{-1ex}
\subsection{Network Representation}
Deep Reinforcement Learning~(DRL) methods usually benefit from the good generalization ability of a deep neural network (DNN), which encodes the state to a vectorized embedding and maps the embedding to the estimated return of the state or action. As the design of representation has a great effect on the efficiency and performance of the algorithm, we will discuss two concerns in the design of the network representation of the SE-state $s_{a_{1:{i}}}^t$ with DNN.

\begin{figure*}[t]
    \centering
    \includegraphics[width=0.9\textwidth]{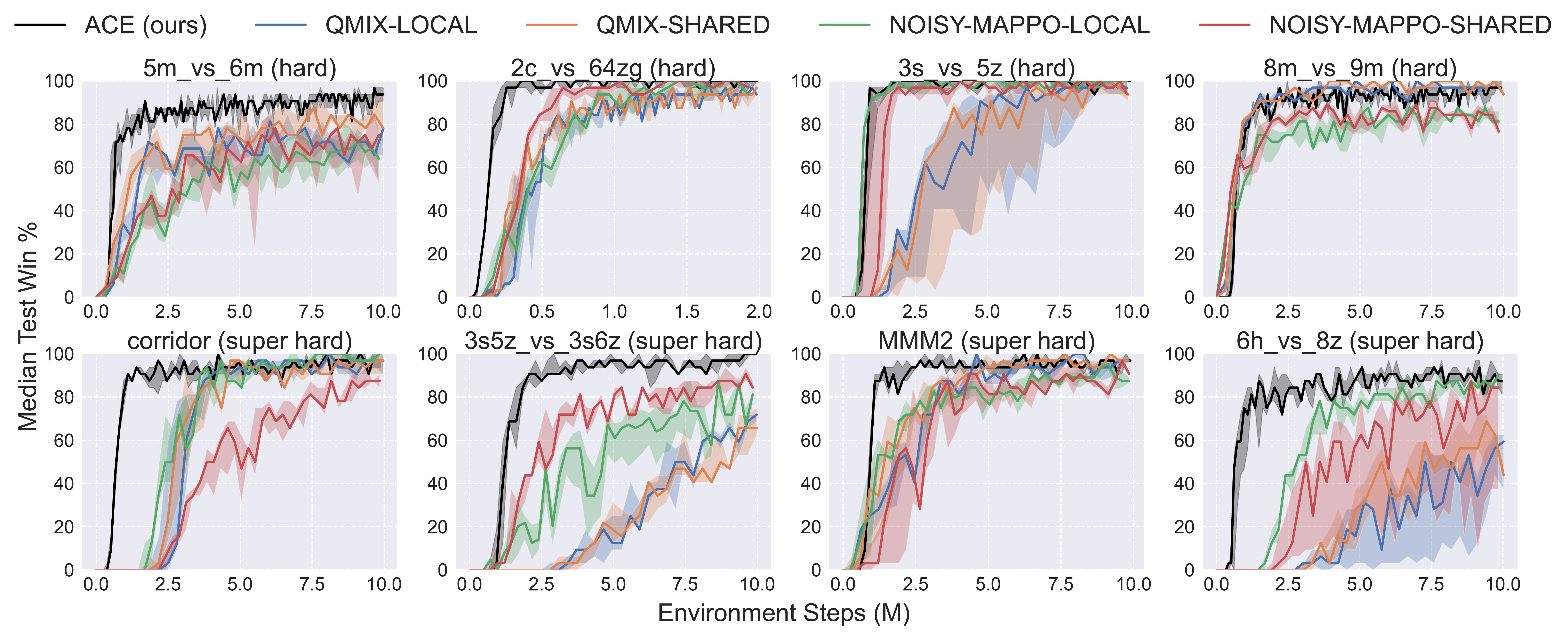}
\vspace{-2ex}
    \caption{Comparison of ACE against baselines on four super hard and four hard SMAC maps.}
\vspace{-4ex}
    \label{fig:smac_main_figure}
\end{figure*}

\textbf{Decomposed State Embedding} \textbf{Firstly}, a transition $\left(s^t,\boldsymbol{a}^t,r\left(s^t,\boldsymbol{a}^t\right),s^{t+1}\right)$ in the original MMDP $\mathcal{G}$ corresponds to $n$ intermediate transitions in the sequential expansion $\widetilde{\mathcal{G}}$ and each intermediate transition requires to evaluate $\left| A_{i} \right|$ next states, resulting in $\sum_{i=1}^n \left| A_{i} \right|$ total states for evaluation. Direct computing all states' embedding from scratch will bring unacceptable computational cost, thus the first principle we follow in the representation of SE-state is: all the state embeddings $e_s\left(s_{a_{1:{i}}}^t\right)$ along the sequential decision-making are decomposed into a shared embedding $e_s\left(s^t\right)$ of the initial state $s^t$, as well as a shared set of embeddings $e_a\left(a_{1}\right),...,e_a\left(a_{n}\right)$ of available actions $a_{1},...,a_{n}$, all generated by the same action encoder. Then, the state embedding $e_s\left(s_{a_{1:{i}}}^t\right)$ is obtained by combining the initial state embedding $e_s\left(s^t\right)$ and the corresponding action embeddings $e_a\left(a_{1}\right),...,e_a\left(a_{i}\right)$. In this decomposition, the original state only requires to be encoded once rather than $\sum_{i=1}^n \left| A_{i} \right|$. Moreover, the combination is additive and introduces negligible cost. \textbf{Secondly}, a multi-agent task involves interaction among multiple units, including cooperative interaction among agent-controlled units, like healing an allied unit in SMAC, and interaction between agent-controlled and environment-controlled units, like attacking an enemy unit in SMAC. We follow two designs, unit-wise state embedding and interaction-aware action embedding, to describe the interactions in the state and action embedding.

\textbf{Unit-wise State Embedding} For state embedding, we use a \textbf{unit encoder} to generate the unit-wise embedding $e_u\left(u_i\right)$ of each unit $u_i$ in the environment, which forms the initial state embedding $e_s\left(s^t\right)=\left[e_u\left(u_1\right),...,e_u\left(u_m\right)\right]$. Here $m$ is the number of units. We assume that the first $n$ units are controlled by the RL agent and the rest $(m-n)$ ones are controlled by the environment. We do not fuse the unit embeddings to a global state embedding, but retain them to facilitate the description of the interactions among units. The input feature of each unit includes the \textbf{node feature} and \textbf{edge feature}. The node feature is the state of each unit, e.g., the health and shield in SMAC and the speed in GRF, and the edge feature is the relation between the units, e.g., the distance between units in SMAC. Our unit encoder takes a fairly simple architecture, depicted in Figure~\ref{fig:unit_encoder}. The node and edge feature are separately encoded by two encoders to generate the corresponding embedding. In this paper, we take a fully connected layer along with a ReLU~\citeyear{relu} as the encoder. The resulted edge embedding average-pooled and then added to the node embedding to obtain the final unit embedding. 

\textbf{Interaction-aware Action Embedding} 
To make the state embedding $e_s\left(s_{a_{1:{i}}}^t\right)$ aware of the unit interactions, we develop a two-fold \textbf{interaction-aware action embedding}. Given an action $a_i$ that is executed by unit $u_i$ and involves the interaction with some target units, its action embedding consists of an \textbf{active embedding} and a \textbf{passive embedding}, formalized by $e_a\left(a_{i}\right) = \left[e_a^{a}\left(a_{i}\right), e_a^{p}\left(a_{i}\right)\right]$. The active embedding $e_a^{a}\left(a_{i}\right)$ encodes the effect of action $a_i$ on the unit $u_i$ itself, and the passive embedding $e_a^p\left(a_{i}\right)$ encodes the effect of action $a_i$ on the target units. For actions without interaction, it only has an active embedding $e_a^a\left(a_{i}\right)$, formalized by $e_a\left(a_{i}\right)=\left[e_a^a\left(a_{i}\right)\right]$.

After the generation of the original state embedding $e_s\left(s^t\right)=\left[e_u\left(u_1\right),...,e_u\left(u_m\right)\right]$ and the action embedding $e_a\left(a_{1}\right),...,e_a\left(a_{i}\right)$, we use an additive combination of the unit and action embedding to construct the state embedding $e\left(s_{a_{1:i}}^t\right)$ of the intermediate SE-states $s_{a_{1:i}}^t$, formalized by $e_s\left(s_{a_{1:i}}^t\right)=\left[e_u\left(u_{1,a_{1:i}}\right),\!...,e_u\left(u_{m,a_{1:i}}\right)\right]$. The element $e_u\left(u_{j,a_{1:i}}\right)$ denotes the combination of the initial unit embedding $e_u\left(u_j\right)$ of unit $j$ and the embeddings of its associated actions among $a_{1:i}$. The rule of combination is: for each action $a_i$, its active action embedding $e_a^a\left(a_i\right)$ is added on the unit embedding $e_u\left(u_i\right)$ of it executor $u_i$; if $a_i$ involves an interaction with some target unit, 
its passive action embedding $e_a^p\left(a_i\right)$ is added on $e_u\left(u_j\right)$ to describe the interaction. The definition of $e_u\left(u_{j,a_{1:i}}\right)$ is formalized by:

\vspace{-3mm}
\begin{small}
\begin{gather}
\label{update}
    e_u\left(u_{j,a_{1:i}}\right)\!=\!
    \begin{cases}
        &e_u\left(u_j\right) + 
        \sum_{e_a^p\left(a_k\right) \in P\left(a_{1:i}\right)_j}
        e_a^p\left(a_k\right),  \\
        & \ \ \ \ \ \ \ \ \ \ \ \ \ \ \ \ \ \ \ \ \ \ \ \ \ \ \ \ \ \ \ \ \ \ \ \ \ \ \ \ \ \ \ \ \ \ \ \ \ \ \ \ \ \ \ \ \ \ \ \text{if $i < j$} \\
        &e_u\left(u_j\right)\!+\!e_a^a\left(a_j\right)\!+\!
        \sum_{e_a^p\left(a_k\right) \in P\left(a_{1:i}\right)_j}
        e_a^p\left(a_k\right), \\
        & \ \ \ \ \ \ \ \ \ \ \ \ \ \ \ \ \ \ \ \ \ \ \ \ \ \ \ \ \ \ \ \ \ \ \ \ \ \ \ \ \ \ \ \ \ \ \ \ \ \ \ \ \ \ \ \ \ \ \ \text{if $i >= j$}
    \end{cases}
\end{gather}
\end{small}
where $P\left(a_{1:i}\right)_j$ is the set of all passive action embeddings whose target unit is $u_j$. When $i>=j$, which means $u_j$ is an agent-controlled unit and has made its decision $a_j$, the active embedding $e_a^a\left(a_j\right)$ will also be added to $e_u\left(u_j\right)$. 

In this paper, the passive embedding $e_a^p\left(a_i\right)$ of a unit $u_i$ is generated from an action encoder whose input is the node feature of the unit $u_i$, because the effect of action $a_i$ may rely on the executor's state. For instance, in GRF the effect on the ball is affected by the speed of the controller. However, the active embedding $e_a^a\left(a_i\right)$ is defined as a learnable parameterized vector, because it is added to the embedding $e_u\left(u_i\right)$ of $u_i$ which has already encoded the state of $u_i$. Both the two kinds of embeddings are learnable. Like the encoders of node and edge features, we also take a fully connected layer along with a ReLU activation as the action encoder in this paper. 

At last, we use a encoder to estimate the value of each SE-state. 
The state embedding $e_s\left(s_{a_{1:i}}^t\right)=\left[e_u\left(u_{1,a_{1:i}}\right),...,e_u\left(u_{m,a_{1:i}}\right)\right]$ is fed into a 'fc-relu' structure to encode the interaction-aware unit embedding, 
followed by a 'pooling-fc' structure to output the estimated value. 
Figure~\ref{fig:transition} demonstrates the pipeline of embeddings generation and how to use them to represent the transition in $\widetilde{\mathcal{G}}$.

\section{Experiment}

To study the advantages of ACE, we consider three tasks: (1) Spiders-and-Fly, (2) StarCraft Multi-Agent Challenge and (3) Google Research Football. Since the baselines we compare with are designed for partial observation settings, we also introduce our efforts to guarantee the fairness in this section. More details on these tasks are included in Appendix.

\textbf{Spiders-and-Fly} The Spiders-and-Fly problem is first proposed in~\citeyear{multiagent_rollout}, where multiple spiders are controlled to catch a fly in a two 2D grid. At each time step, each spider moves to a neighboring location or stays put, while the fly moves randomly to a neighboring location. In this paper, we modify it to a much harder problem where only two spiders are controlled by the RL agent, and the fly will avoid moving to the neighboring locations of the spiders, otherwise stay still. Each episode starts with a state where the Manhattan distance between the fly and each spider is larger than 4. With such modifications, the two spiders must perform perfect cooperation to encircle the fly at the corner. The reward is defined as 10 if the fly is caught otherwise 0. 

\textbf{StarCraft Multi-Agent Challenge~(SMAC)}
In SMAC~\citeyear{smac}, the ally units controlled by the RL agent play against the enemy units controlled by the built-in rules. To win the competition, allies learn to perform cooperative micro-tricks, such as positioning, kiting and focusing fire. This benchmark consists of various maps classified as Easy, Hard, and Super Hard. Since the Easy maps solved well by existing methods~\citeyear{pymarl2}, we focus on four super hard maps: corridor, MMM2, 6h\_vs\_8z, and 3s5z\_vs\_3s6z, and four hard maps: 5m\_vs\_6m, 2c\_vs\_64zg, 8m\_vs\_9m and 3s\_vs\_5z. 

\textbf{Google Research Football~(GRF)}
Compared with SMAC, GRF~\citeyear{grf} provides a harder environment with large action space and sparse reward. 
In the GRF, agents coordinate timing and location to organize attacks and only scoring leads to rewards. In our experiments, we control the left team players except for the goalkeeper while the built-in engine controls the right team players. 
We evaluate our method on two challenging scenarios: academy\_3\_vs\_1\_with\_keeper and academy\_counterattack\_hard.
For a fair comparison, we use the standard 19 actions (\textit{i.e.}, moving, sliding, shooting and passing), and use the same observation in CDS~\citeyear{cds} to construct our input feature. Following the settings in CDS, we also make a reasonable change to the two half-court offensive scenarios: we will lose if our players or the ball returns to our half-court. All experiments are tested with this modification. 
The final reward is +100 when our team wins, -1 when our player or the ball returns to our half-court, and 0 otherwise.

\textbf{Evaluation Metric}
For Spiders-and-Fly, we derive an analytical optimal solution as the oracle policy and introduce two metrics: (1) the samples required to achieve 100\% success rate in ten steps, and (2) the gap between the average steps required by the RL policy and the oracle policy to catch the fly.
For SMAC, we follow the official evaluation metric in~\citeyear{smac}, \textit{i.e.,} we run 32 test episodes without exploration to record the test win rate and report the median performance as well as the 25-75\% percentiles across 5 seeds. For GRF, we similarly run 32 test episodes to obtain win rate and report the average win rate as well as the variance across 5 seeds. 

\vspace{-1ex}
\begin{table}[t!]
    \centering
    \resizebox{0.99\linewidth}{!}{
    \begin{tabular}{c|c|cccc}
        \toprule
        Metric &Map &VDN &QMIX &QTRAN &ACE\\
        \midrule
        \multirow{2}{*}{Steps}      & 5$\times$5   &0.78$\pm$0.10 &0.77$\pm$0.10 &0.60$\pm$0.09 &\textbf{0.04$\pm$0.03}	   \\
                                    & 7$\times$7   &0.90$\pm$0.12 &0.87$\pm$0.11 &1.02$\pm$0.09 &\textbf{0.07$\pm$0.02}	   \\
        \midrule
        \multirow{2}{*}{Samples (M)}  & 5$\times$5   &0.19$\pm$0.02 &0.19$\pm$0.02 &0.17$\pm$0.02 &\textbf{0.09$\pm$0.01}	   \\
                                    & 7$\times$7   &1.97$\pm$0.10 &1.81$\pm$0.09 &1.68$\pm$0.09 &\textbf{1.01$\pm$0.06}	   \\
        \bottomrule
    \end{tabular}}
    \vspace{-1ex}
    \caption{Comparison ACE against baselines on Spiders-and-Fly. Steps represent the gap between the average steps of the methods and the oracle policy. Samples represent the number of samples to achieve a 100\% success rate within 10 steps.}
    \label{tab:spider_fly}
\vspace{-4ex}
\end{table}

\subsection{Performance}
\vspace{-1ex}

\textbf{Spiders-and-Fly} 
We compare ACE with three value factorization methods: QTRAN~\citeyear{qtran}, QMIX~\citeyear{qmix} and VDN~\citeyear{vdn}, on two grids with the sizes 5x5 and 7x7. 
As shown in Table~\ref{tab:spider_fly}, ACE is the only one that can approximate the performance of the oracle policy, while the baselines, although also find the best behavior in some cases, cannot consistently converge to the optimal policy. 
Moreover, ACE takes up to 50\% fewer samples to achieve the 100\% success rate in ten steps.

\begin{figure}[t]
    \centering
    \includegraphics[width=0.95\linewidth]{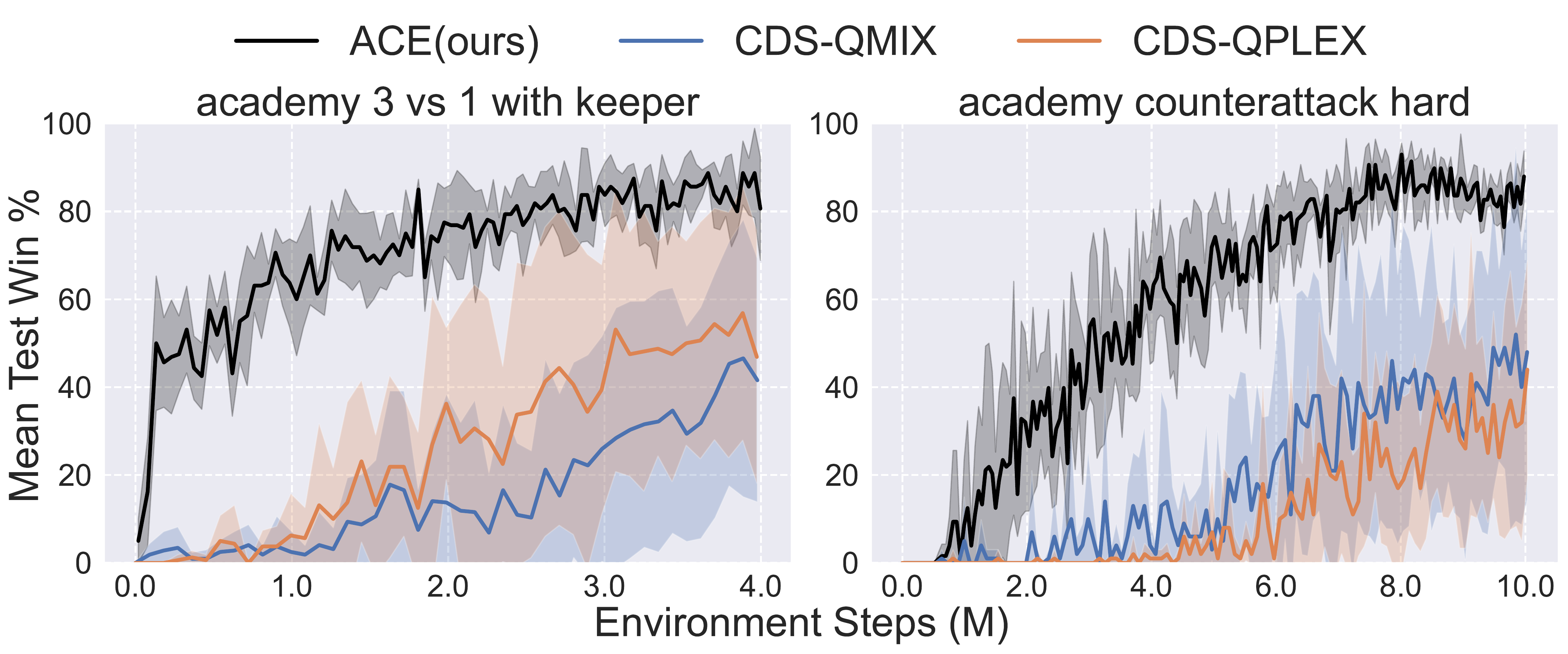}
\vspace{-2ex}
    \caption{Comparison of ACE against baseline on GRF.
}
\label{fig:grf_main_figure}
\vspace{-4ex}
\end{figure}

\textbf{SMAC}
We compare ACE with both the state-of-the-art (SOTA) value-based and actor-critic methods on SMAC. First, our value-based baseline is the fine-tuned QMIX~\citeyear{pymarl2} combining QMIX~\citeyear{qmix} with bags of code-level optimizations and outperforming QPLEX~\citeyear{wang2020qplex}, QTRAN~\citeyear{qtran}, vanilla QMIX and Weighted QMIX~\citeyear{wqmix}. Secondly, we choose the SOTA actor-critic method, NOISY-MAPPO~\citeyear{noisy_mappo}, as the actor-critic baseline. Although the two methods are proposed for CTDE pipeline, they are also important baselines to solve the exponentially large action space in multi-agent tasks. To this end, the comparison with them is fair.
Note that both baseline algorithms are designed for partially observable scenarios, where each agent only use its local observation to generate the action, while ACE uses the observation of all units to make decisions. Thus, to make a fair comparison, in the two baselines, we make each agent share the union of all units' observations at the input, denoted as SHARED. We also evaluate the baselines with the original local observation, denoted as LOCAL, because in some cases the shared observation has worse performance. For example, NOISY-MAPPO-LOCAL achieves better performance than NOISY-MAPPO-SHARED in 6h\_vs\_5z. As shown in Figure~\ref{fig:smac_main_figure}, ACE surpasses fine-tuned QMIX and NOISY-MAPPO by a large margin in the final win rate and the sample efficiency. Remarkably, it achieves 100\% test win rates in almost all maps, including 5m\_vs\_6m and 3s5z\_vs\_3s6z, which have not been solved well by existing methods even with shared observation. Therefore, ACE achieves a new SOTA on SMAC.

\textbf{GRF}
We show the performance comparison against the baselines in Figure~\ref{fig:grf_main_figure}. ACE outperforms the SOTA methods CDS-QMIX~\citeyear{cds} and CDS-QPLEX~\citeyear{cds} by a large margin in both two scenarios. The gap between ACE and the baselines is even larger than that on SMAC, possibly due to that the football game requires more complex cooperation skills.
\vspace{-2ex}

\begin{figure*}[h]
    \begin{center}
    \resizebox{0.9\linewidth}{!}{
    \subfloat[]{
        \begin{minipage}{0.5\textwidth}
        \includegraphics[width=1.0\linewidth]{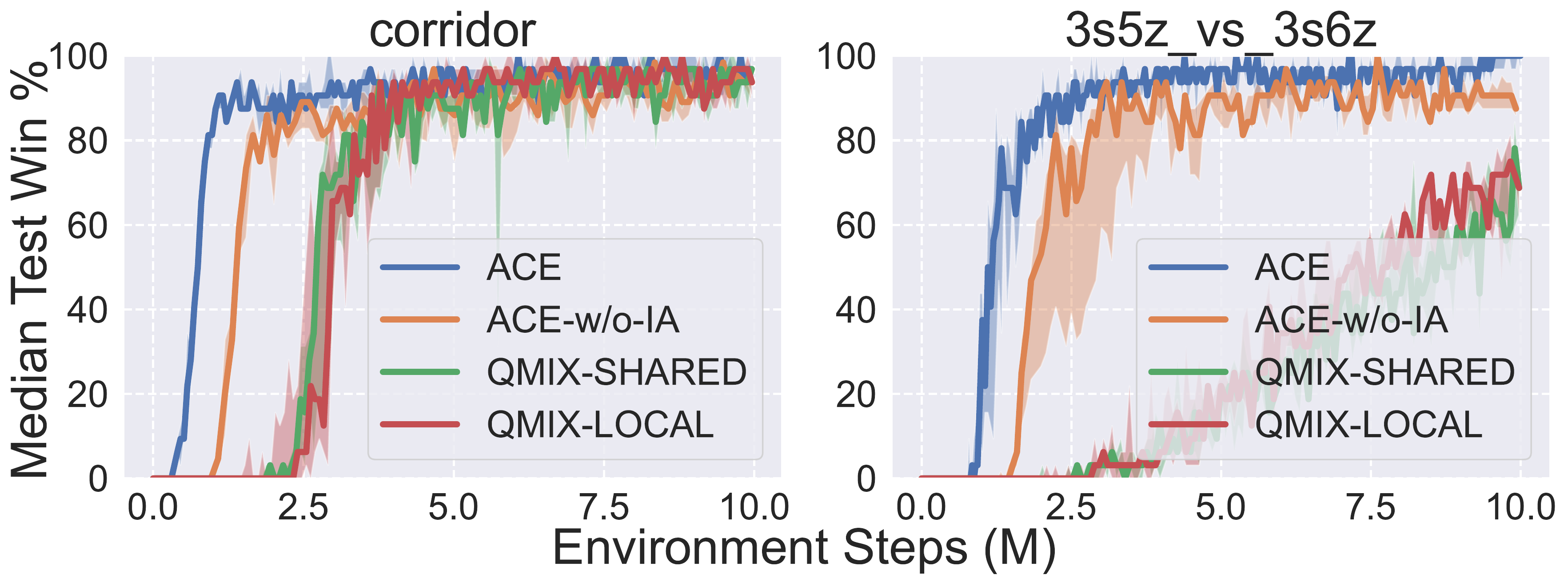}
        \vspace{-8pt}
        \label{fig:smac_vdn_figure}
        \end{minipage}
        \vspace{-8pt}
        }
    \subfloat[]{
        \begin{minipage}{0.5\textwidth}
        \includegraphics[width=1.0\linewidth]{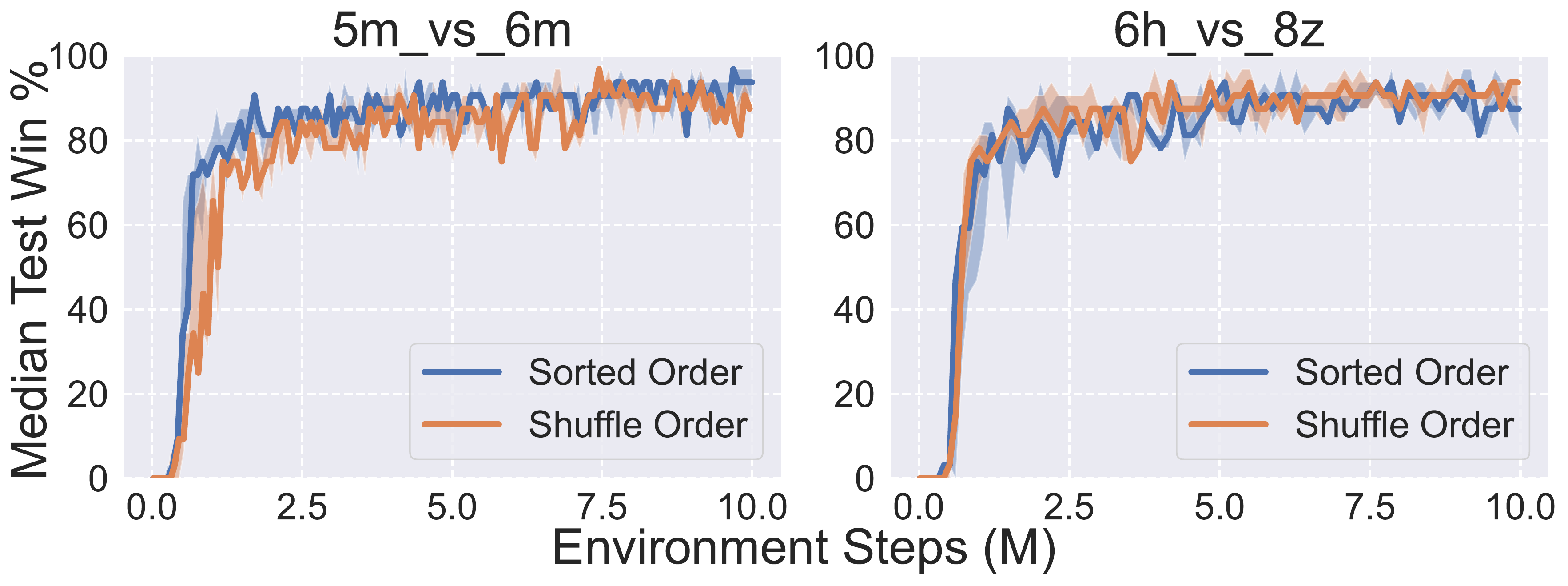}
        \label{fig:smac_order_figure}
        \vspace{-8pt}
        \end{minipage}
        \vspace{-8pt}
    }
    }
    \end{center}
\vspace{-12pt}
    \caption{\textbf{(a):} Comparison of ACE and ACE-w/o-IA against QMIX on corridor and 3s5z\_vs\_3s6z. \textbf{(b):} Comparison of sorted order against shuffle order of ACE.}
\vspace{-3ex}
\end{figure*}

\begin{figure*}[t]
\centering
    \resizebox{0.9\linewidth}{!}{
    \subfloat[]{
    \begin{minipage}{0.25\textwidth}
    \includegraphics[width=\linewidth]{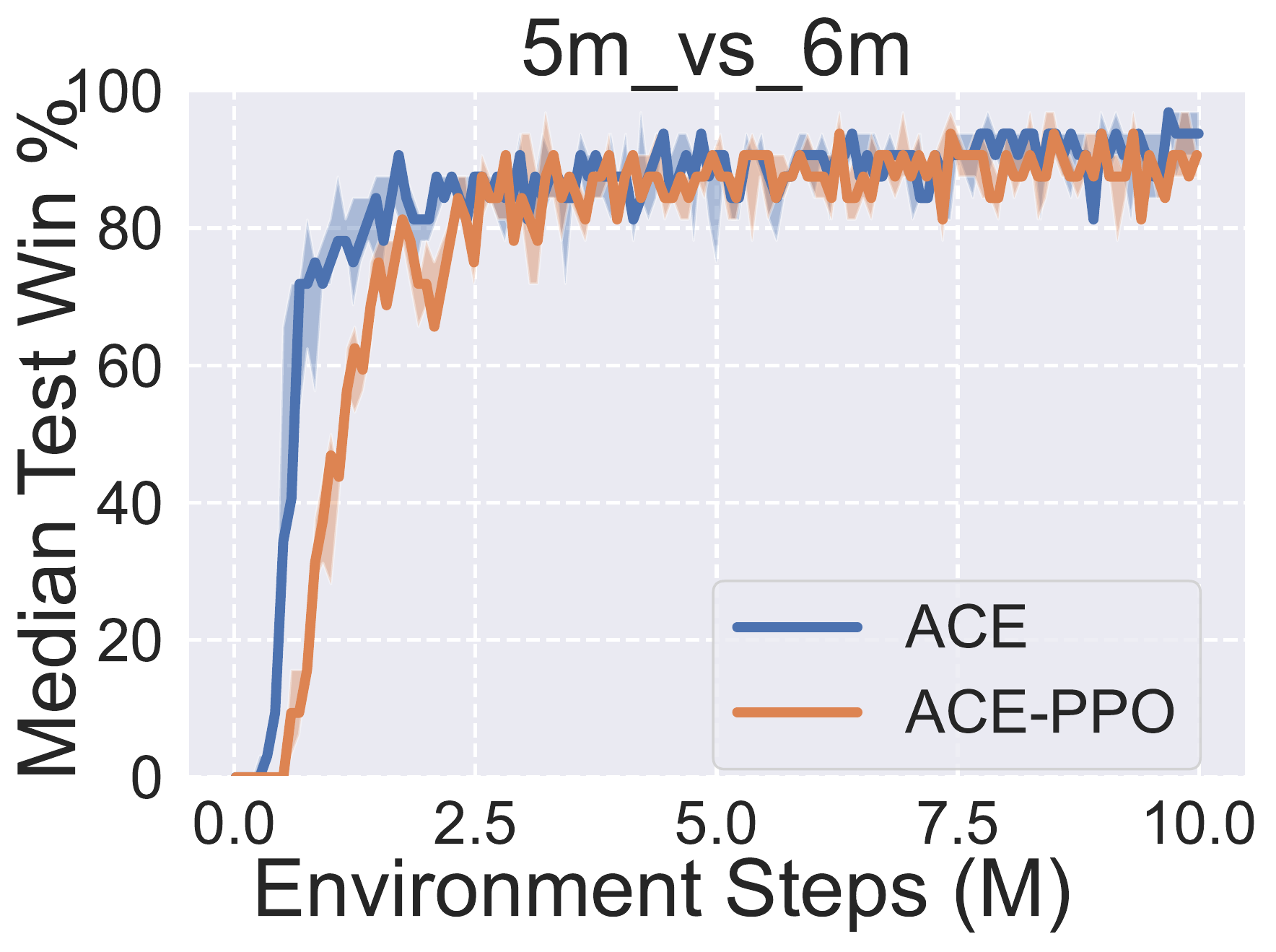}
    \label{fig:smac_dqn_ppo_figure}
    \vspace{-16pt}
    \end{minipage}
    }
    \subfloat[]{
    \begin{minipage}{0.25\textwidth}
    \includegraphics[width=\linewidth]{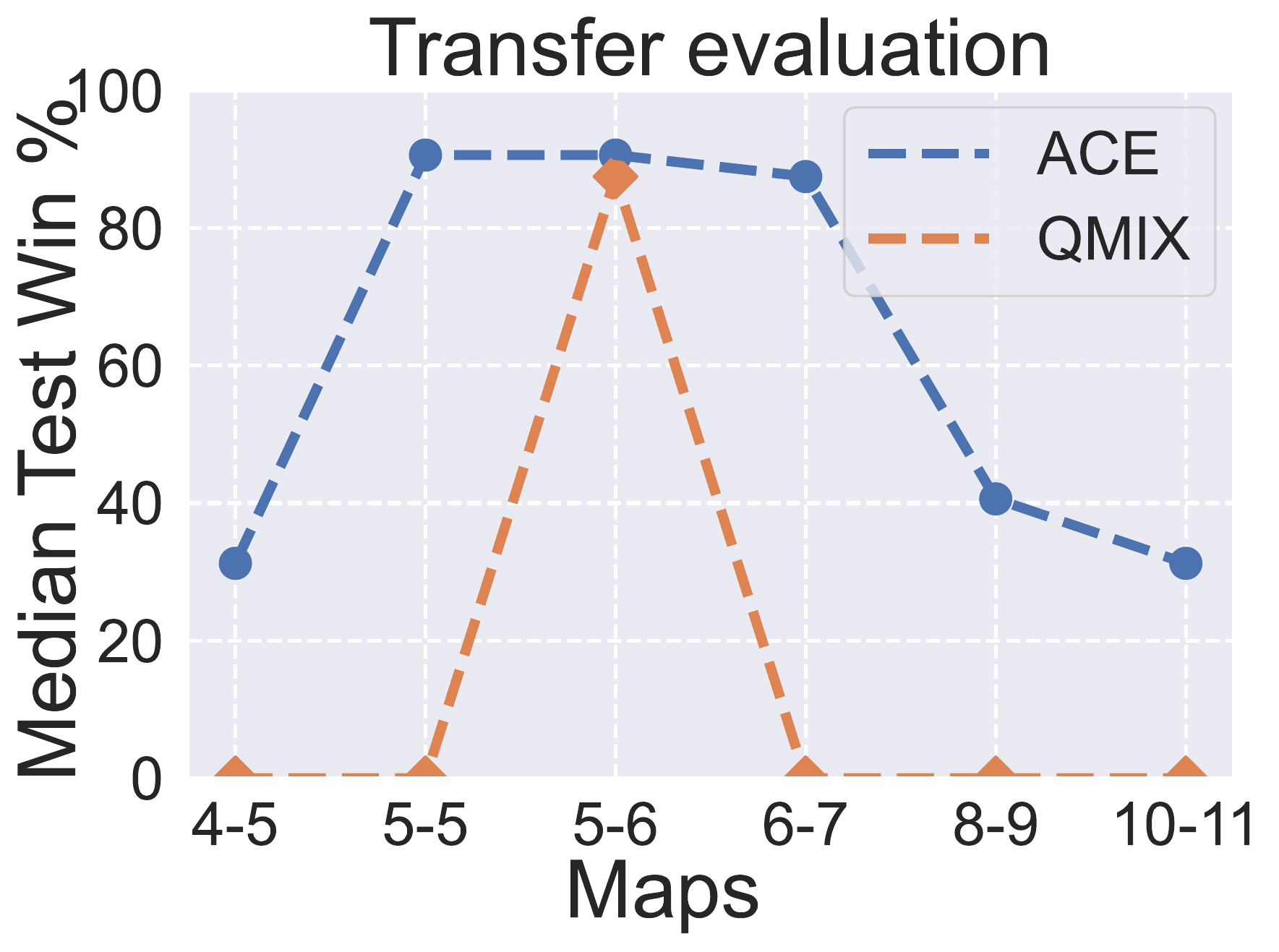}
    \label{fig:transfer}
     \vspace{-8pt}
    \end{minipage}
    \vspace{-8pt}
    }
    \subfloat[]{
    \begin{minipage}{0.5\textwidth}
    \includegraphics[width=\linewidth]{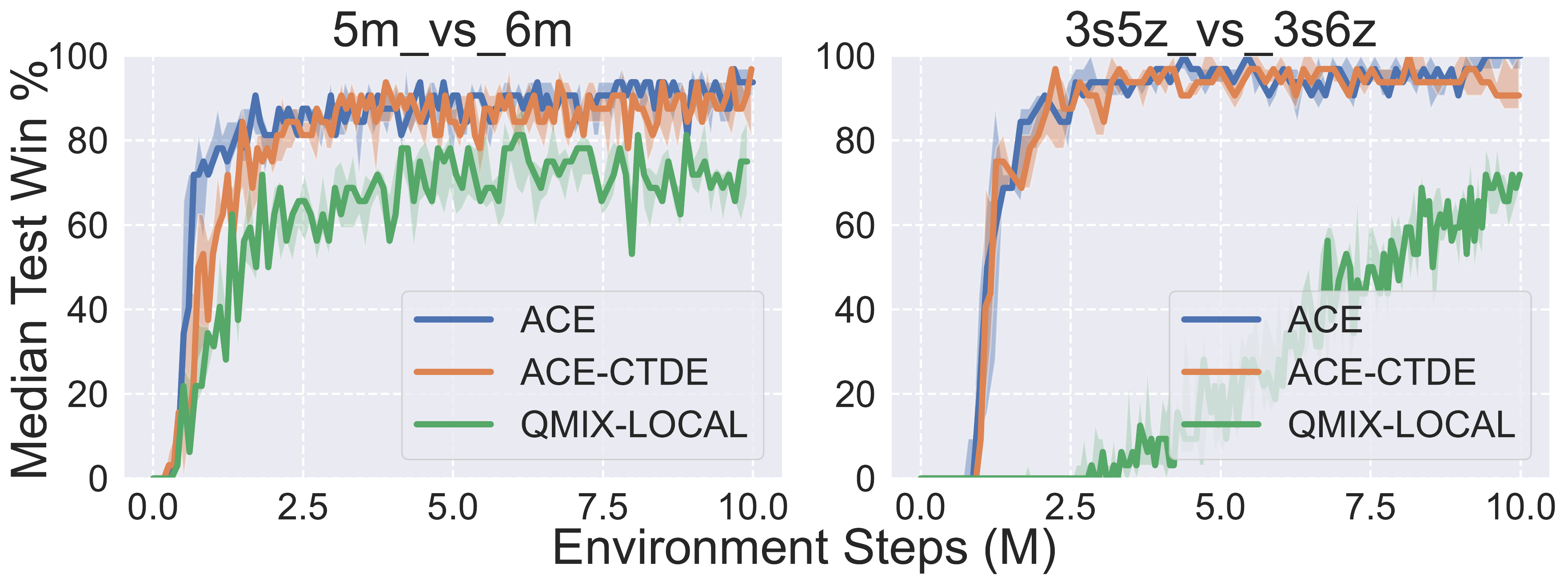}
    \label{fig:smac_de_figure}
    \vspace{-8pt}
    \end{minipage}
    \vspace{-8pt}
    }
    }
\vspace{-8pt}
    \caption{\textbf{(a):} Comparision of ACE and ACE-PPO on 5m\_vs\_6m.
    \textbf{(b):} Transfer from 5m\_vs\_6m to maps of different numbers of agents. 'x-y' represents xm\_vs\_ym. ACE can achieve remarkable performance in the zero-shot setting, regardless of whether agent number increases or decreases. 
    \textbf{(c):} Comparison of ACE-CTDE against ACE.
}
\vspace{-4ex}
\end{figure*}

\vspace{1ex}
\subsection{\textbf{Ablation:} What matters in the components of ACE?}
\vspace{-1ex}
To better understand why ACE outperforms the baseline algorithms, we further make ablations and modifications to it. First, we remove the interaction-aware action embedding by only using the active embedding, denoted by ACE-w/o-IA, and compare it with ACE and the fine-tuned QMIX. As shown in Figure~\ref{fig:smac_vdn_figure}, the gap between ACE-w/o-IA and the QMIX is still large, which validates that the bidirectional action-dependency itself retains much of the benefit of ACE. Moreover, ACE-w/o-IA is worse than ACE due to the effective interaction-aware embedding. Secondly, we study how the order of decision-making influences the performance of ACE. We compare two settings: 1) Shuffle Order: random orders are generated for both data collection and training. 2) Sorted Order: agents are sorted by unit types and locations. As shown in Figure~\ref{fig:smac_order_figure}, the two settings have little difference in performance in two SMAC maps, which validates that ACE is quite robust to the order of agents.
\vspace{-1ex}
\subsection{\textbf{Extension:} Extend ACE to the actor-critic method}
\vspace{-1ex}
Our approach, transforming a MMDP into a MDP, is general and can be combined with a more extensive range of single-agent RL algorithms. In this section, we combine ACE with an actor-critic method, PPO, denoted by ACE-PPO. To generate the logit of each action in PPO, we roll out each action to the corresponding next SE-states, and use the same way how our value encoder evaluates these states to obtain the logit. 
As shown in Figure~\ref{fig:smac_dqn_ppo_figure}, ACE-PPO achieves a comparable performance with ACE on the 5m\_vs\_6m map, which validates that ACE is applicable to wider types of algorithms.

\vspace{-1ex}
\subsection{\textbf{Generalization:} Does ACE generalize to a new task with a different number of agents?}
\vspace{-1ex}

An interesting advantage of ACE is its surprising generalization. Compared with prior methods where agents make decisions individually, ACE explicitly models the cooperation between agents. As a result, when the preceding agents take sub-optimal actions due to the change of the task, the subsequent agents compensate it through the learned cooperative skills. 
We train ACE and the fine-tuned QMIX-SHARED on 5m\_vs\_6m and test them on 4m\_vs\_5m, 5m\_vs\_5m, 6m\_vs\_7m, 8m\_vs\_9m and 10m\_vs\_11m. 
Note that the action and observation space change with the agent number, we address this problem as described in Appendix.
As in Figure~\ref{fig:transfer}, although without any fine-tuning on the test maps, ACE still achieves considerable win rates, which reveals an excellent generalization to the change of the agent number. 

\vspace{-1ex}
\subsection{\textbf{Practicability:} Apply ACE to the CTDE scheme.}
\vspace{-1ex}
We develop a simple adaptation of ACE, denoted by ACE-CTDE, to apply it in the Centralized training and decentralized execution (CTDE) scheme, a popular scheme for multi-agent tasks with limited communication. In most CTDE methods, an individual value function uses the local observation to estimate the individual value, and a joint value function estimates the value of the joint action with the global state. The optimal actions of the two functions are aligned via well-designed constraints to guarantee the IGM property.
Similarly, we use a counterfactual distillation technique, to distill the optimal joint action directly generated via the sequential rollout in ACE, into an additional individual value function $Q\left(o_i^t,a_i\right)$ of which the input is the local observation. The counterfactual distillation is formalized by $
    \hat{Q}\left(o_i^t,a_i\right)=
    V\left(s_{a_i,a_{i-}^*}^t\right)$.
$\hat{Q}\left(o_i^t,a_i\right)$ is the target to update $Q\left(o_i^t,a_i\right)$ and $o_i^t$ is the local observation of agent $i$. $a_{i-}^*$ denotes the optimal joint action generated by the sequential rollout excluding the action of agent $i$. This distillation estimates each individual action value of an agent with other agents' actions fixed jointly optimal, thus it follows the IGM principle. 
In Figure~\ref{fig:smac_de_figure}, ACE-CTDE is evaluated with the individual value function $Q$ in a decentralized way. We can see that ACE-CTDE performs nearly as well as ACE due to the IGM property of the proposed distillation.

\section{Conclusion}
In this paper, we introduce bidirectional action-dependency to solve the non-stationary problem in cooperative multi-agent tasks. The proposed ACE algorithm significantly improves the sample efficiency and the converged performance against the state-of-the-art algorithm. Comprehensive experiments validate the advantages of ACE in many aspects. 

\section{Acknowledgments}
Wanli Ouyang was supported by the Australian Research Council Grant DP200103223, Australian Medical Research Future Fund MRFAI000085, CRC-P Smart Material Recovery Facility (SMRF) – Curby Soft Plastics, and CRC-P ARIA - Bionic Visual-Spatial Prosthesis for the Blind.
This work is partially supported by the Shanghai Committee of Science and Technology (Grant No. 21DZ1100100).
We acknowledge Yining Fang for the strong support and in depth discussion.

\clearpage
\bibliography{aaai23.bib}
\clearpage
\clearpage
\section{Appendix}
In this appendix, we first discuss the difference between this paper and related works. Then, we provide more elaboration on the implementation details, experiment results and qualitative results.
Specifically, we present the discussion of the related works in \nameref{sec: rlt_discussion}, the implementation details of ACE and the baselines on the three environments in \nameref{sec: details},
implementation details of ablation studies in \nameref{sec: ablation},
and additional experimental results in \nameref{sec: add_exp}.

\section{Further Discussion of Related Work}
\label{sec: rlt_discussion}

Many MARL algorithms have been proposed to achieve improvements in cooperation aspects. The two tracks that are most relative with ACE are coordination graph~\citeyear{zhang2011coordinated, zhang2013coordinating, bohmer2020deep} and advanced network representation~\citeyear{wang2019action}.

To mitigate the non-stationary problem in MARL, the coordination Graph-based algorithms introduce action dependency into the formulation and utilize the higher-order state-action functions instead of fully individual state-action functions to represent the joint Q-function. However, coordination graph methods, still based on the value factorization paradigm, only introduce pairwise action dependency in the value estimation for MMDP. While ACE is based on the transformation from MMDP to MDP, which tackles the non-stationary problem by introducing full action dependency. 

To boost the action representation in MARL, ASN~\citeyear{wang2019action} also classifies the actions of an agent into $A_{in}$ and $A_{out}$ according to whether the actions affect other units. However, ACE is fundamentally different with ASN. First, ASN is designed for individual value network, while ACE for the joint value network. Secondly, ASN can only model the action of one agent, and ACE describe all actions that has been selected. Moreover, ASN uses the information of the target unit at the input to model the action that interacts with the target unit, while ACE use the action embedding to update the output embedding of the target unit.

ZO~\citeyear{usunier2016episodic} is another relative work. It also introduces a sequential inference scheme/sequential expansion (SE) for structured output prediction. However, the potential of SE has long been overlooked by the community. In this work, we rethink this topic and fully releases the potential of SE and further validate the compatibility of SE with other single agent algorithms (\textit{e.g.,} PPO). Additionally, ACE directly learns the value function by $V(s,a_1,a_2,...,a_i)$, which has a unified representation of agents that have and haven't made decisions. While the definition of value function in ZO is $Q(a_i|s,a_1,a_2,...,a_{i-1})$, where the model must additionally learn which agent to choose act for in each step, which struggles to achieve high sample efficiency.

\section{Implementation Details}
\label{sec: details}
\subsection{Baselines}
\paragraph{(1) Fine-tuned QMIX\citeyear{pymarl2} and Noisy-MAPPO\citeyear{noisy_mappo}}
We use the official code of \href{https://github.com/hijkzzz/pymarl2}{fine-tuned QMIX} \footnote{https://github.com/hijkzzz/pymarl2} and \href{https://github.com/hijkzzz/noisy-mappo}{Noisy-MAPPO} \footnote{https://github.com/hijkzzz/noisy-mappo} provided by their authors. These two implementations incorporate fine-tuned hyper-parameters and bags of code-level optimizations, which fully release the potential of the two baselines.

\paragraph{(2) The Difference between SHARED and LOCAL} The aim of the SHARED setting is to enable each individual agent in our baseline algorithms to make the decision based on the observation of all agents. In the SHARED setting, we make the state of an enemy observable to all agents if it is observed by at least one agent. Also, the state of each agent is observed by all other agents. As a result, the SHARED setting enables all agents to have access to a shared global observation, which is the union of the observation of all agents.

\paragraph{(3) CDS\citeyear{cds}} 
We use the official code of \href{https://github.com/lich14/CDS}{CDS} \footnote{https://github.com/lich14/CDS}.

\paragraph{(4) QTRAN, QMIX, and VDN on Spider and Fly~\citeyear{cds}}
We use the same hyper-parameters as ACE for QTRAN, QMIX, and VDN, as listed in Table~\ref{appendix:spider_hyperparameter}. Also, the input feature of them is based on the same information as ACE, including the unit ID, the position, and the distance among units, as listed in Table~\ref{appendix:spider_feature}. For the algorithm-specific hyper-parameters of these baselines, like the weight of the extra loss in QTRAN, we use the original choices provided in their original papers.
\begin{table}[h]
\centering
\resizebox{0.68\linewidth}{!}{
\begin{tabular}{cc}
\toprule
\multicolumn{1}{c|}{Parameter}        & Value                            \\ \midrule
\multicolumn{2}{c}{Exploration}                                       \\ \midrule
\multicolumn{1}{c|}{action\_selector}      & epsilon\_greedy                  \\
\multicolumn{1}{c|}{epsilon\_type}         & linear                           \\
\multicolumn{1}{c|}{epsilon\_start}        & 1                                \\
\multicolumn{1}{c|}{epsilon\_end}          & 0.05                             \\
\multicolumn{1}{c|}{epsilon\_decay}        & 150k             \\ 
\midrule
\multicolumn{2}{c}{Sample}                                           \\ \midrule
\multicolumn{1}{c|}{collector\_env\_num}   & 8                                \\
\multicolumn{1}{c|}{sample\_per\_collect}            & 1024                               \\
\multicolumn{1}{c|}{replay\_buffer\_size}  & 1M                             \\ \midrule
\multicolumn{2}{c}{Training}                                          \\ \midrule
\multicolumn{1}{c|}{update\_per\_collect}  & 10                               \\
\multicolumn{1}{c|}{batch\_size}           & 256                              \\
\multicolumn{1}{c|}{weight\_decay}         & 0                         \\
\multicolumn{1}{c|}{learning\_rate}        & 0.0005
\\
\multicolumn{1}{c|}{target\_update\_theta} & 0.02                            \\
\multicolumn{1}{c|}{discount\_factor}      & 0.99                             \\
\multicolumn{1}{c|}{optimizer}             & adam                          \\ \midrule
\multicolumn{2}{c}{Model}                                             \\ \midrule
\multicolumn{1}{c|}{hidden\_len}           & 128                              \\
\bottomrule
\end{tabular}
}
\vspace{-0.2cm}
\caption{Hyper-parameter Settings of ACE and baselines on Spiders and Fly.}
\vspace{-3ex}
\label{appendix:spider_hyperparameter}
\end{table}

\begin{table}[h]
\centering
\resizebox{0.9\linewidth}{!}{
\begin{tabular}{cc}
\toprule
\multicolumn{1}{c|}{Feature Name}        & Components                            \\ \midrule
\multicolumn{2}{c}{ACE}             \\ 
\midrule
\multicolumn{1}{c|}{node\_feature}      & unit id (0, 1 for spiders, 2 for fly) \\
\multicolumn{1}{c|}{}                   & unit position               \\
\multicolumn{1}{c|}{edge\_feature}         & distance on two axes \\
\midrule
\multicolumn{2}{c}{Baselines}       \\
\midrule
\multicolumn{1}{c|}{input\_feature}         & unit id (0, 1 for spiders, 2 for fly) \\
\multicolumn{1}{c|}{}                       & unit position, distance to all other\\
\multicolumn{1}{c|}{}                       &  units on two axes \\
\bottomrule
\end{tabular}}
\vspace{-0.2cm}
\caption{Input Feature on Spiders and Fly.}
\vspace{-3ex}
\label{appendix:spider_feature}
\end{table}

\subsection{ACE}
We implemented our ACE based on the \href{https://github.com/opendilab/DI-engine}{DI-engine} \footnote{https://github.com/opendilab/DI-engine}, which is a generalized decision intelligence engine and supports various deep reinforcement learning algorithms. All codes are licensed under the Apache License or MIT License.
The hyper-parameter settings of ACE on Spiders and Fly, SMAC and GRF are respectively listed in Table~\ref{appendix:spider_hyperparameter}, Table~\ref{appendix:smac_hyperparameter} and \ref{appendix:grf_hyperparameter}. The definition of the input feature on the three environments are provided in Table~\ref{appendix:spider_feature}, Table~\ref{appendix:smac_feature} and \ref{appendix:grf_feature}.
Different from \href{https://github.com/hijkzzz/pymarl2}{pymarl2} \footnote{https://github.com/hijkzzz/pymarl2}, in DI-engine, 
the batch\_size represents the number of transitions, not the episode.

\begin{table}[h]
\centering
\resizebox{0.8\linewidth}{!}{
\begin{tabular}{cc}
\toprule
\multicolumn{1}{c|}{Parameter}        & Value                            \\ \midrule
\multicolumn{2}{c}{Exploration}                                       \\ \midrule
\multicolumn{1}{c|}{action\_selector}      & epsilon\_greedy                  \\
\multicolumn{1}{c|}{epsilon\_type}         & linear                           \\
\multicolumn{1}{c|}{epsilon\_start}        & 1                                \\
\multicolumn{1}{c|}{epsilon\_end}          & 0.05                             \\
\multicolumn{1}{c|}{epsilon\_decay}        & 50k                            \\ \midrule
\multicolumn{2}{c}{Sampler}                                           \\ \midrule
\multicolumn{1}{c|}{collector\_env\_num}   & 8                                \\
\multicolumn{1}{c|}{episode\_per\_collect}            & 32                               \\
\multicolumn{1}{c|}{replay\_buffer\_size}  & 300k (100k for 2c\_vs\_64zg) \\ \midrule
\multicolumn{2}{c}{Training}                                          \\ \midrule
\multicolumn{1}{c|}{update\_per\_collect}  & 50                               \\
\multicolumn{1}{c|}{batch\_size}           & 320                              \\
\multicolumn{1}{c|}{weight\_decay}         & 1e-5                         \\
\multicolumn{1}{c|}{learning\_rate}        & 3-4                         \\
\multicolumn{1}{c|}{target\_update\_theta} & 0.008                            \\
\multicolumn{1}{c|}{discount\_factor}      & 0.99                             \\
\multicolumn{1}{c|}{optimizer}             & rmsprop                          \\ \midrule
\multicolumn{2}{c}{Model}                                             \\ \midrule
\multicolumn{1}{c|}{hidden\_len}           & 256                              \\ \bottomrule
\end{tabular}
}
\vspace{-0.2cm}
\caption{Hyperparameter Settings of ACE on SMAC.}
\vspace{-3ex}
\label{appendix:smac_hyperparameter}
\end{table}

\begin{table}[h]
\centering
\resizebox{0.99\linewidth}{!}{
\begin{tabular}{cc}
\toprule
\multicolumn{1}{c|}{Feature Name}        & Components                            \\ \midrule
\multicolumn{1}{c|}{node\_feature}      & unit id, unit type, unit position, health, shield,                  \\
\multicolumn{1}{c|}{}                   &cool down\\
\multicolumn{1}{c|}{edge\_feature}         & distance on two axes                \\
\bottomrule
\end{tabular}
}
\vspace{-0.2cm}
\caption{Input Feature of ACE on SMAC.}
\vspace{-3ex}
\label{appendix:smac_feature}
\end{table}

\begin{table}[h]
\centering
\resizebox{0.95\linewidth}{!}{
\begin{tabular}{cc}
\toprule
\multicolumn{1}{c|}{Feature Name}        & Components                            \\ \midrule
\multicolumn{1}{c|}{node\_feature}      & unit id, unit position, speed, whether own ball  \\
\multicolumn{1}{c|}{edge\_feature}         & distance on two axes                           \\
\bottomrule
\end{tabular}
}
\vspace{-0.2cm}
\caption{Input Feature of ACE on GRF.}
\vspace{-3ex}
\label{appendix:grf_feature}
\end{table}

In Table~\ref{appendix: embedding}, we show in detail how the two-fold interaction-aware action embedding was constructed.

\begin{table*}[t]
\centering
\resizebox{0.9\linewidth}{!}{
\begin{tabular}{c|c|cc|cc}
\toprule
\multirow{2}{*}{Env}             & \multirow{2}{*}{Action}  & \multicolumn{2}{c}{Passive Embedding} & \multicolumn{2}{c}{Active Embedding} \\ \cline{3-6} 
                                 &                          & Whether Use?           & Target Unit              & Whether Use?             & Target Unit          \\ \midrule
\multirow{2}{*}{Spiders-and-Fly} & move                     & No         & N/A                      & Yes           & Itself               \\
                                 & stay                     & No         & N/A                      & Yes           & Itself               \\ \midrule
\multirow{3}{*}{SMAC}            & move                     & No         & N/A                      & Yes           & Itself               \\
                                 & attack                   & Yes        & The unit it attack       & Yes           & Itself               \\
                                 & heal                    & Yes        & The unit it heal        & Yes           & Itself               \\ \midrule
\multirow{2}{*}{GRF}             & action of the ball owner & Yes        & No                       & Yes           & Itself               \\
                                 & action of other players  & No         & N/A                      & Yes           & Itself               \\ \bottomrule
\end{tabular}}
\vspace{-0.2cm}
\caption{Interaction-aware action embedding.}
\vspace{-3ex}
\label{appendix: embedding}
\end{table*}

\subsection{Environments}
\subsubsection{Spiders and Fly}

\begin{figure}[h]
    \centering
    \includegraphics[width=0.35\textwidth]{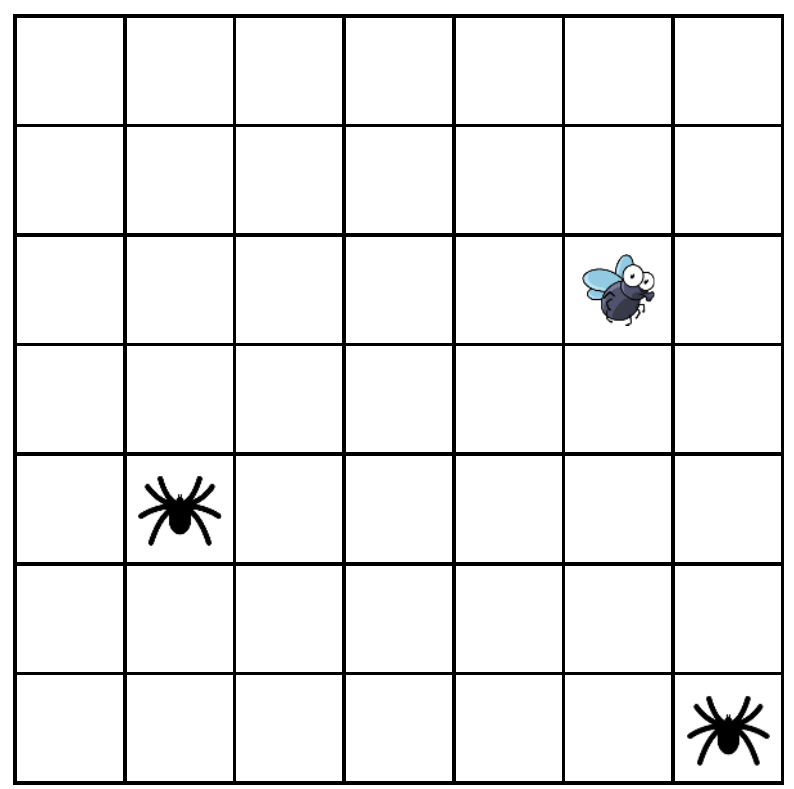}
\vspace{-0.2cm}
    \caption{Visualization of spiders and the fly in a 7$\times$7 map, where the two spiders are controlled by the RL agent. The reward is defined as 10 if the fly is caught otherwise 0.
}
\vspace{-3ex}
    \label{appendix:spides-fly}
\end{figure}

Here, we visualize the initial positions of the two spiders and fly in a 7$\times$7 map in the Figure~\ref{appendix:spides-fly}. Each episode starts with a state where the Manhattan distance between the fly and each spider is larger than 4. In each time step, the optimal strategy is one of the following two types: (1) both of the spiders move to drive the fly to the corner, and when it is at the corner, (2) one spider stays still to restrict the possible movement of the fly and another one approaches the fly. Thus, optimal cooperation is required in each time step. It is observed in our experiments that, the proposed ACE is the only one that can approximate the performance of the oracle policy in the 5$\times$5 and 7$\times$7 map. This result benefits from that the introduced bidirectional action-dependency enables the explicit learning of how to cooperate with the preceding agents and how the successors react.

\subsubsection{GRF}
\begin{figure}[t]
  \centering
\subfloat[academy\_3\_vs\_1\_with\_keeper]{
\begin{minipage}{0.45\textwidth}    %
    \includegraphics[width=1.0\linewidth]{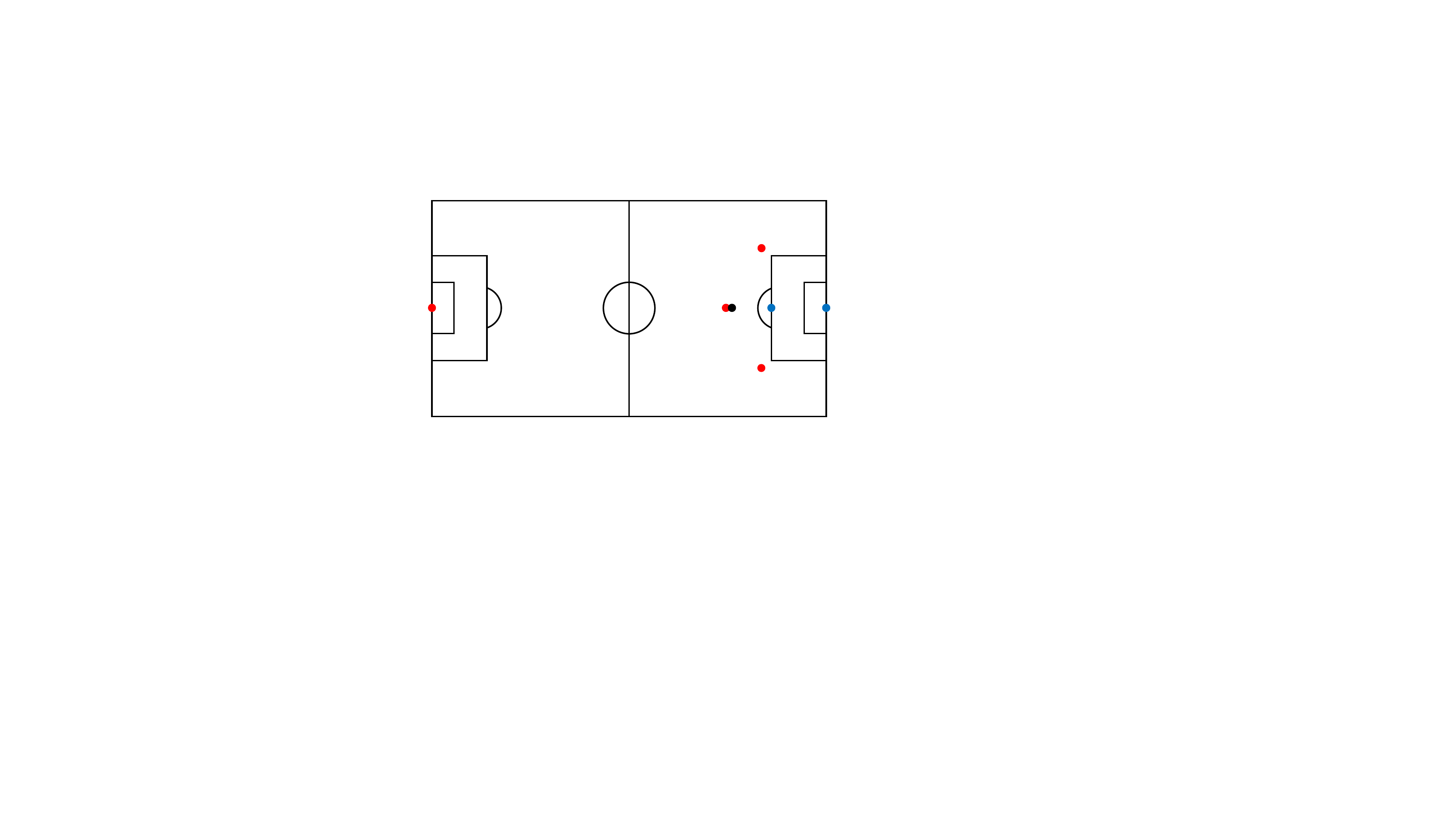}
\end{minipage}
}
\hspace{3ex}
\subfloat[academy\_counterattack\_hard]{
\begin{minipage}{0.45\textwidth}
    \includegraphics[width=1.0\linewidth]{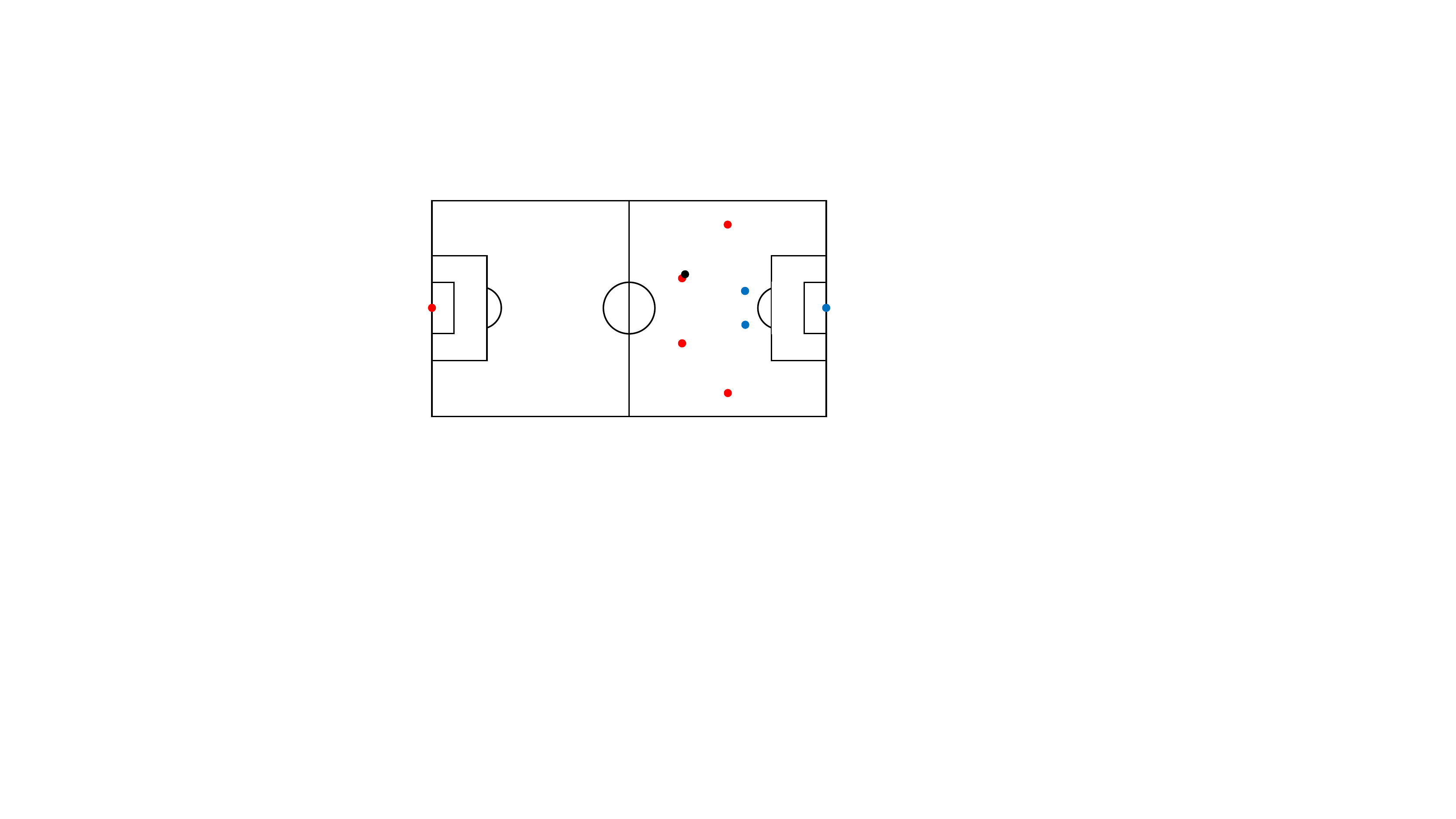}
\end{minipage}
}
  \caption{Visualization of the initial positions of all players and the ball, where red points are our players, blue points are opponents and the black point represents the ball. All our players except for our goalkeeper are controlled by an RL agent while others are controlled by the built-in engine.}
 \label{appendix:grf_vis}
 \vspace{-3ex}
\end{figure}

\begin{figure*}[h]
    \centering
    \includegraphics[width=0.84\textwidth]{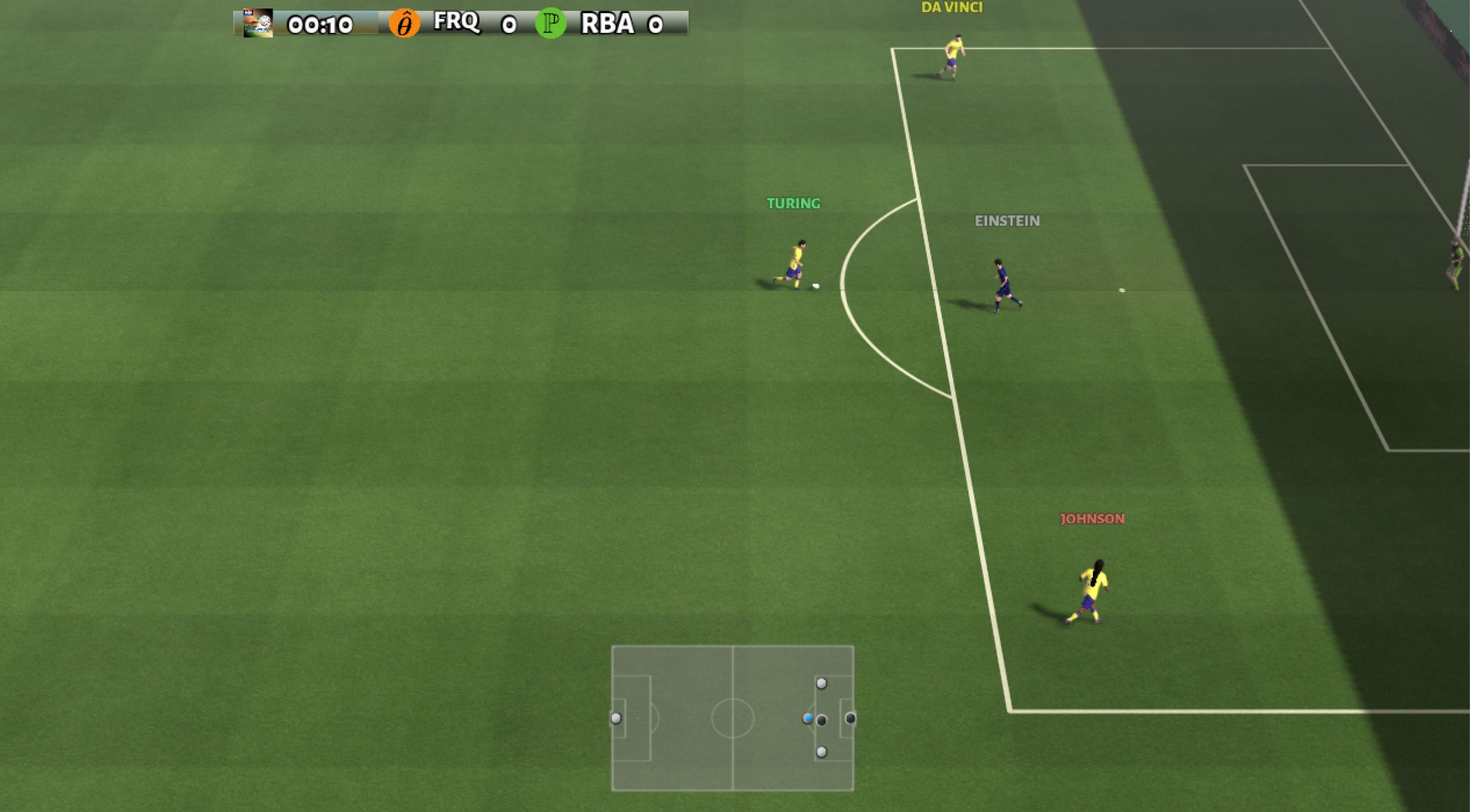}
\vspace{-0.2cm}
    \caption{The academy\_3\_vs\_1\_with\_keeper scenario in real game.
}
\vspace{-3ex}
    \label{appendix:grf}
\end{figure*}

Academy\_3\_vs\_1\_with\_keeper and academy\_counter-
attack\_hard are two of the hardest official scenarios in GRF. To provide more clear details about the two scenarios in GRF, we visualize the initial positions of all players and the ball in the two scenarios in the Figure~\ref{appendix:grf_vis}, and provide an RGB screenshot of academy\_3\_vs\_1\_with\_keeper in Figure~\ref{appendix:grf}. In both two scenarios, the target is to control the left team players (red points in Figure~\ref{appendix:grf_vis}) to get a score with the learned cooperative skills. The right team players are controlled by a built-in AI to defend. The proposed ACE achieves a new SOTA in both two scenarios.

\section{Ablation Details}
\label{sec: ablation}
\subsection{ACE-PPO}
In ACE-PPO, we derive the probability distribution over the action space of each agent via a softmax operation over a logit vector, formalized by:
\begin{gather}
\label{ppo_prob}
   p\left(a_{i+1}^t \mid s_{a_{1:i}}^t\right)=
   \frac{e^{l\left(a_{i+1}^t \mid s_{a_{1:i}}^t\right)}}{\sum_{\tilde{a}_{i+1}^t \in \mathcal{A}_{i+1}}e^{l\left(\tilde{a}_{i+1}^t \mid s_{a_{1:i}}^t\right)}}.
\end{gather}
Here $l\left(a_{i+1}^t \mid s_{a_{1:i}}^t\right)$ is the logit generated in the same way as $V\left(s_{a_{1:i+1}}^t\right)$. Specifically, we use an additional logit encoder with the same structure as the value encoder to produce $l\left(a_{i+1}^t \mid s_{a_{1:i}}^t\right)$.

We use GAE($\lambda$) as the advantage estimator, which is a common practice of PPO in most tasks. The temporal difference is calculated and summed over all adjacent SE-states, formalized by:
\begin{gather}
\label{ppo_gae}
   A\left(a_{i+1}^t \mid s_{a_{1:i}}^t\right) = \\ 
   (\gamma\lambda)^0 \left(\gamma V\left(s_{a_{1:i+1}}^t\right) - V\left(s_{a_{1:i}}^t\right)\right) \\
   +(\gamma\lambda)^1 \left(\gamma V\left(s_{a_{1:i+2}}^t\right) - V\left(s_{a_{1:i+1}}^t\right)\right) \\
   +\ldots\\
   +(\gamma\lambda)^{(n-i)} \left(r\left(s^{t}, a_{1:n}^t\right)+\gamma V\left(s_{a_{1}}^{t+1}\right) - V\left(s_{a_{1:n}}^t\right)\right) \\
   +(\gamma\lambda)^{(n-i+1)} \left(\gamma V\left(s_{a_{2}}^{t+1}\right) - V\left(s_{a_{1}}^{t+1}\right)\right) \\
   +\ldots.
\end{gather}

In summary, our implementation of ACE-PPO on the SE-MDP is equivalent to the standard PPO on a single agent MDP. The hyper-parameters are listed in Table~\ref{appendix:ppo_smac_hyperparameter}.
\begin{table}[h!]
\centering
\resizebox{0.52\linewidth}{!}{
\begin{tabular}{cc}
\toprule
\multicolumn{1}{c|}{Parameter}        & Value                            \\ \midrule
\multicolumn{2}{c}{Exploration}                                       \\ \midrule
\multicolumn{1}{c|}{action\_selector}      & softmax                 \\ \midrule
\multicolumn{2}{c}{Sampler}                                           \\ \midrule
\multicolumn{1}{c|}{collector\_env\_num}   & 8                                \\
\multicolumn{1}{c|}{n\_episode}            & 32                               \\
\multicolumn{1}{c|}{gae\_lambda}  & 0.95 \\ \midrule
\multicolumn{2}{c}{Training}                                          \\ \midrule
\multicolumn{1}{c|}{update\_per\_collect}  & 50                               \\
\multicolumn{1}{c|}{batch\_size}           & 3200                              \\
\multicolumn{1}{c|}{learning\_rate}        & 5e-4  \\
\multicolumn{1}{c|}{discount\_factor}      & 0.99                             \\
\multicolumn{1}{c|}{optimizer}             & adam                          \\ 
\multicolumn{1}{c|}{value\_weight}      & 1    \\
\multicolumn{1}{c|}{entropy\_weight}    & 0.01 \\
\multicolumn{1}{c|}{clip\_ratio}        & 0.05 \\
\multicolumn{1}{c|}{use\_value\_clip}   & True \\
\multicolumn{1}{c|}{value\_clip\_ratio} & 0.3  \\
\multicolumn{1}{c|}{recompute\_adv}     & True \\
\multicolumn{1}{c|}{adv\_norm}          & True \\
\multicolumn{1}{c|}{value\_norm}        & True \\
\midrule
\multicolumn{2}{c}{Model}                                             \\ \midrule
\multicolumn{1}{c|}{hidden\_len}           & 256                              \\
\bottomrule
\end{tabular}
}
\caption{Hyperparameter Settings of ACE-PPO on SMAC.}
\vspace{-3ex}
\label{appendix:ppo_smac_hyperparameter}
\end{table}

\subsection{Generalization}
To handle the change of the observation dimension and the action space among maps with different agent numbers, we develop extended observation and action space. In all the maps, we assume that there exist the largest possible numbers of ally units and enemy units, which are 10 and 11 in the largest map 10m\_vs\_11m. To realize any smaller map xm\_vs\_ym(with x<=10 and y<=11), we initialize each episode with only x ally units and y enemy units alive, which are randomly chosen. This modification enables us to run ACE and fine-tuned QMIX on all maps with the same observation dimension and action space.

\begin{figure*}[t]
\centering
    \subfloat[]{
    \begin{minipage}{0.5\textwidth}
    \includegraphics[width=\linewidth]{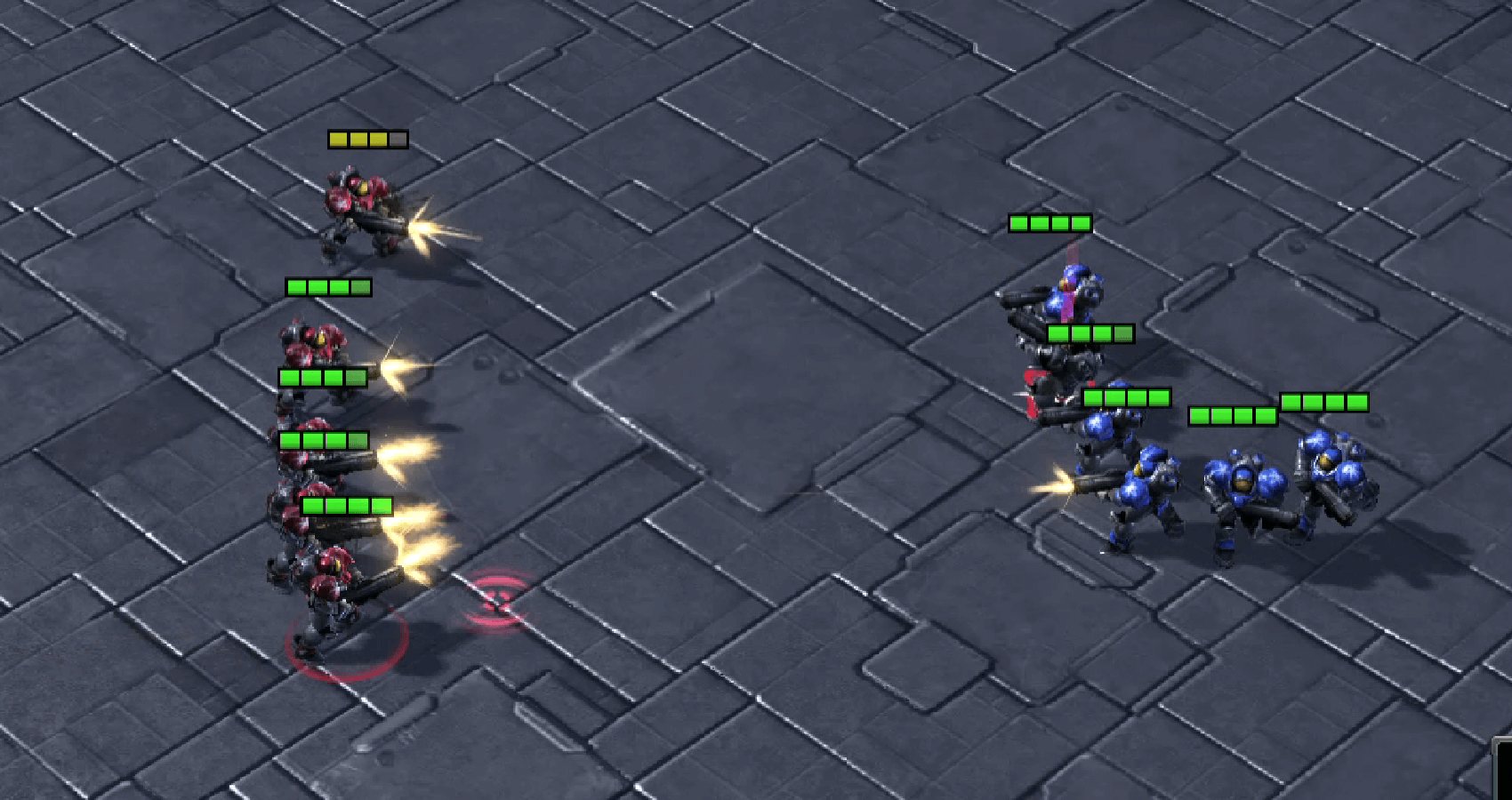}
    \label{appendix:5m6m_position}
    \vspace{-16pt}
    \end{minipage}
    }
    \subfloat[]{
    \begin{minipage}{0.5\textwidth}
    \includegraphics[width=\linewidth]{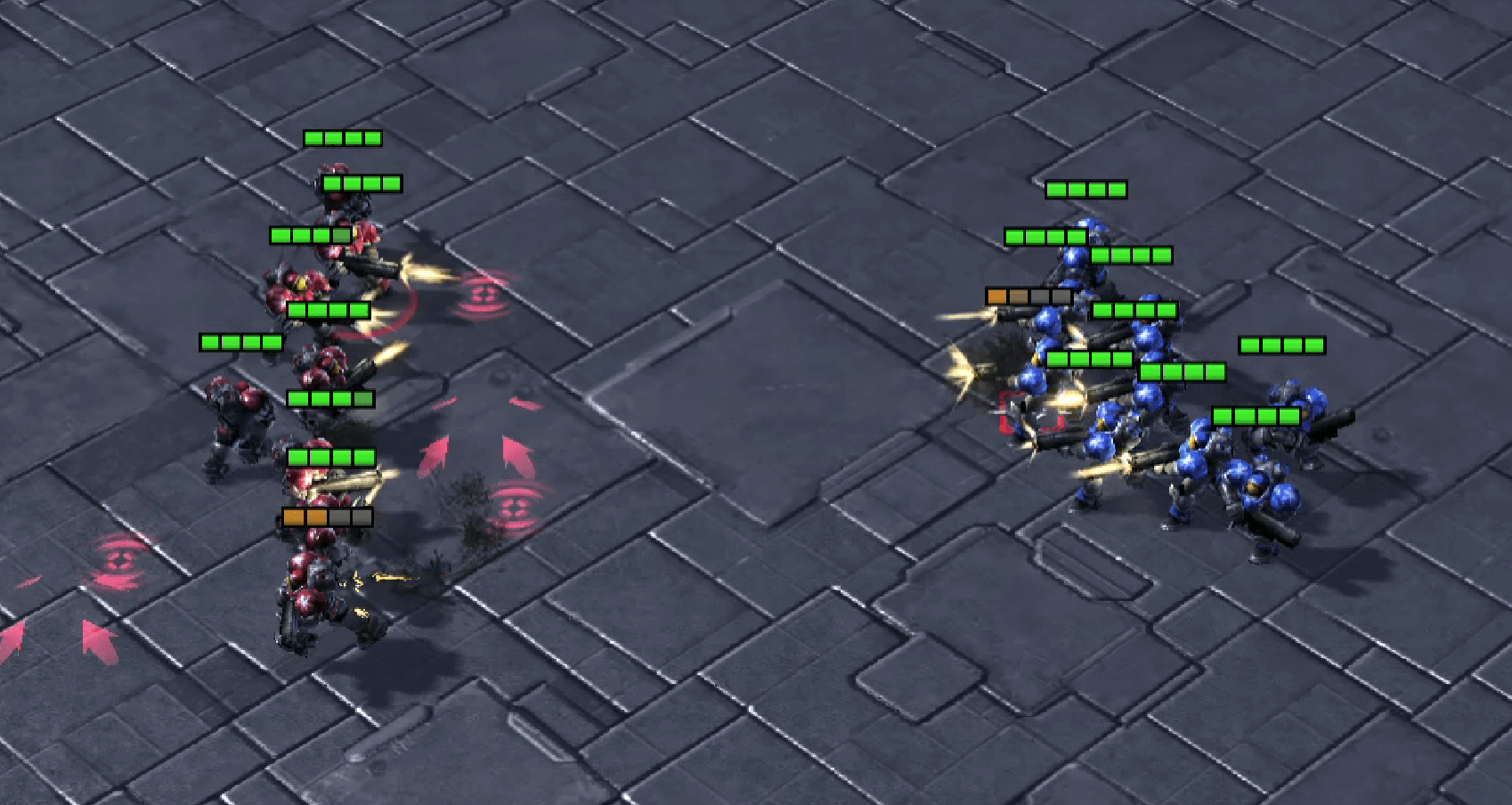}
    \label{appendix:5m6mto8m9m_position}
     \vspace{-8pt}
    \end{minipage}
    \vspace{-8pt}
    }
    \quad
    \subfloat[]{
    \begin{minipage}{0.5\textwidth}
    \includegraphics[width=\linewidth]{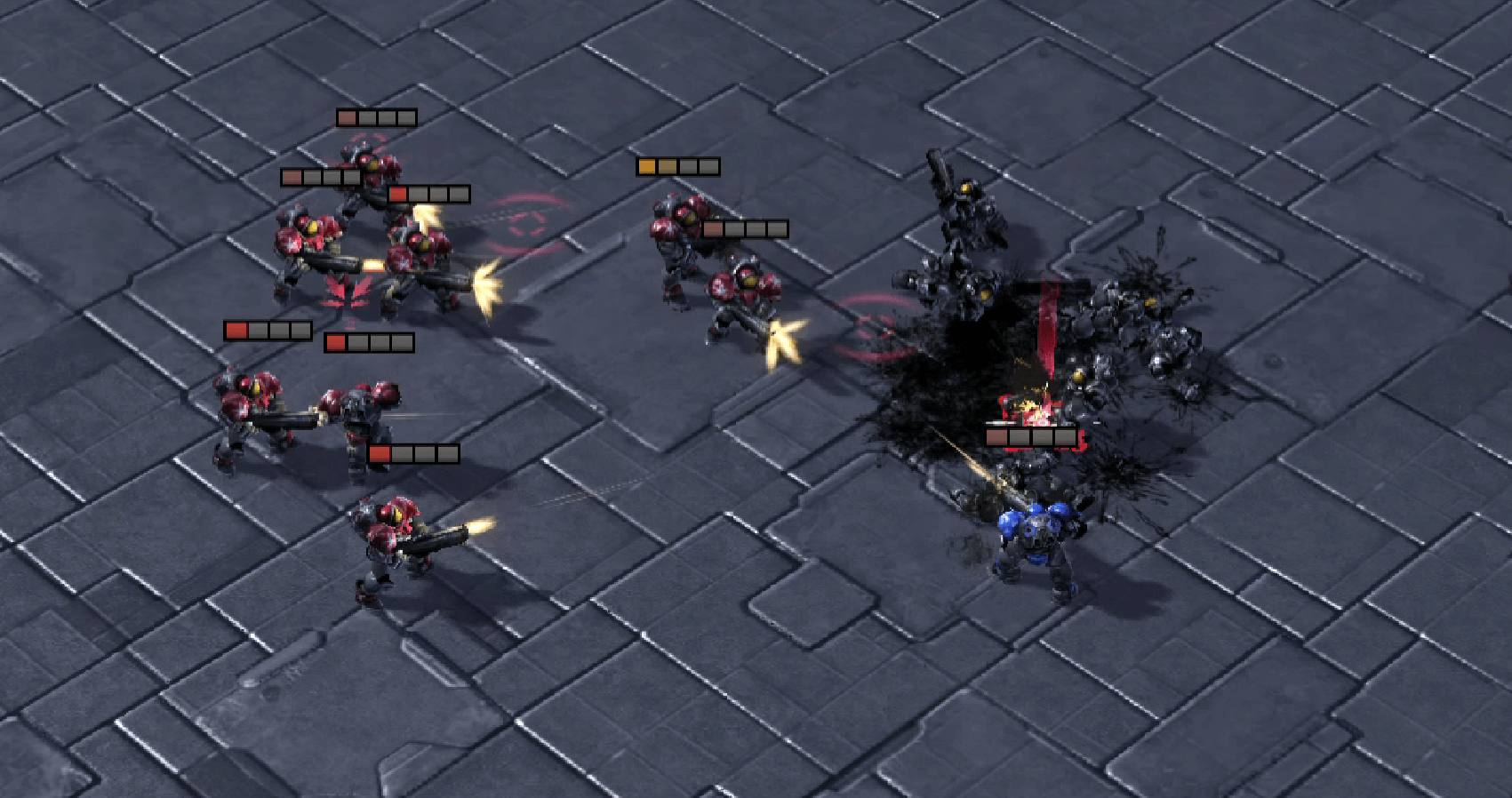}
    \label{appendix:5m6mto8m9m_zouwei}
    \vspace{-8pt}
    \end{minipage}
    \vspace{-8pt}
    }
    \subfloat[]{
    \begin{minipage}{0.5\textwidth}
    \includegraphics[width=\linewidth]{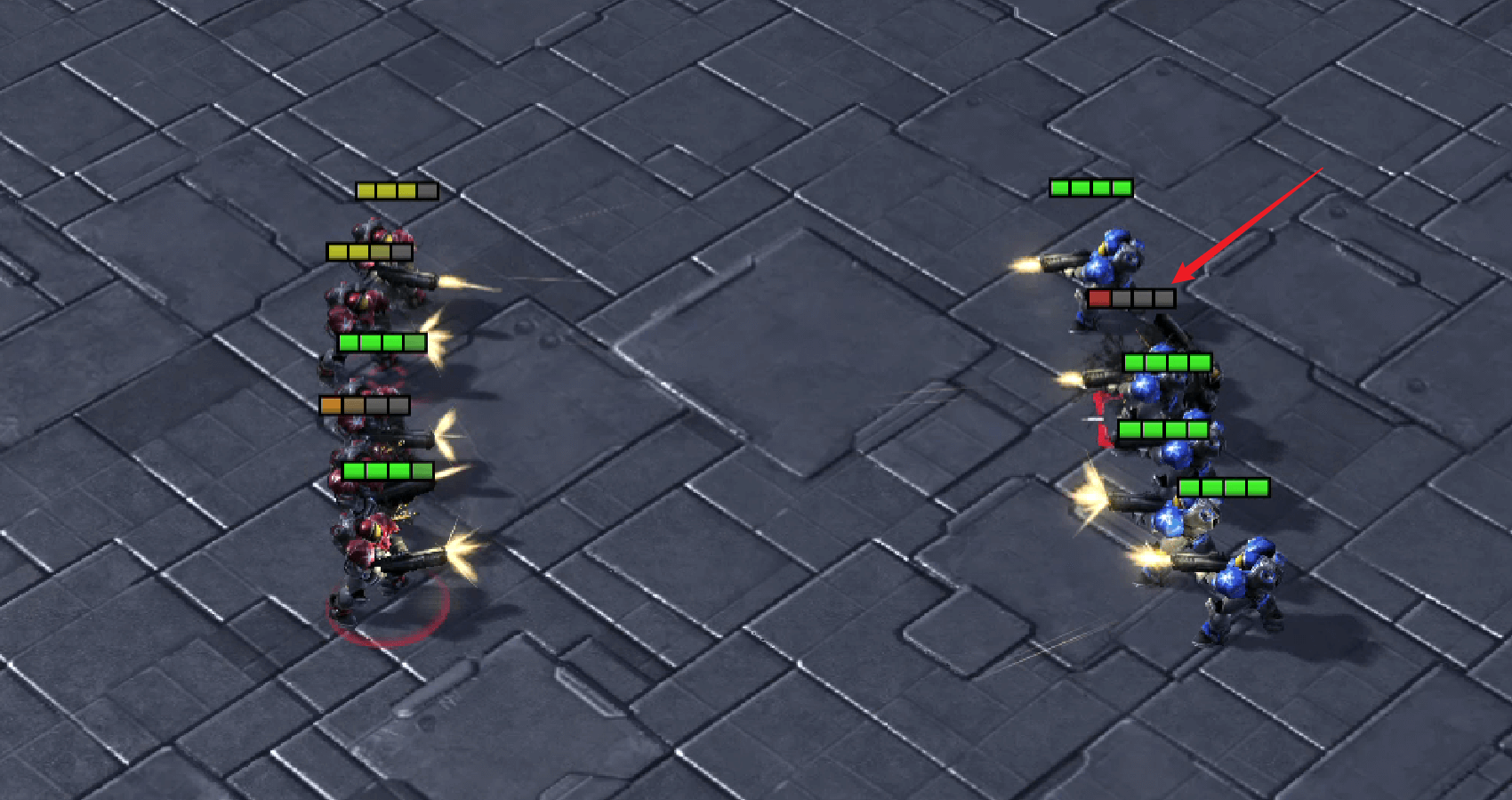}
    \label{appendix:5m6m_fire}
     \vspace{-8pt}
    \end{minipage}
    \vspace{-8pt}
    }
    \quad
    \subfloat[]{
    \begin{minipage}{0.5\textwidth}
    \includegraphics[width=\linewidth]{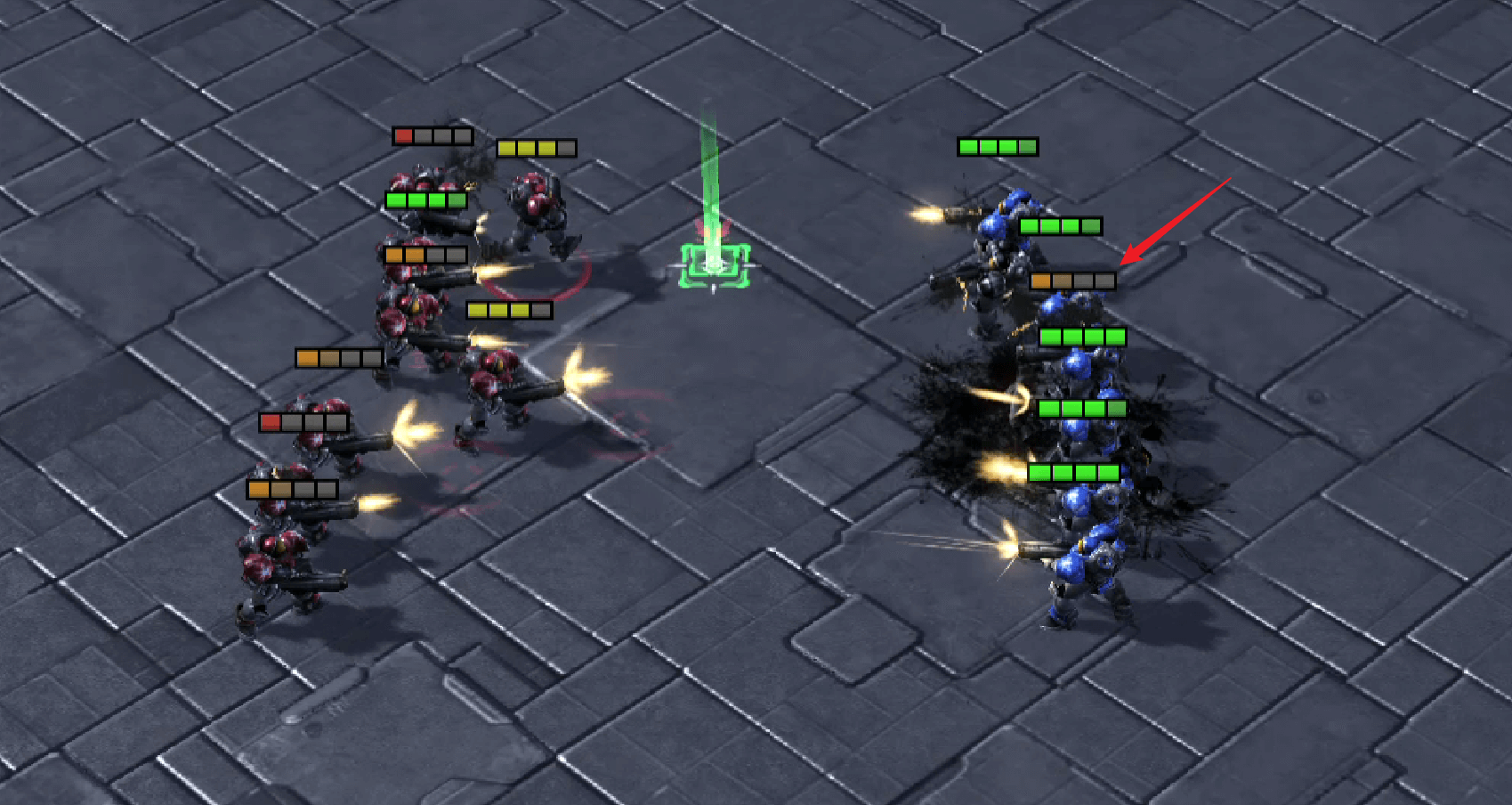}
    \label{appendix:5m6mto8m9m_fire}
     \vspace{-8pt}
    \end{minipage}
    \vspace{-8pt}
    }
    \subfloat[]{
    \begin{minipage}{0.5\textwidth}
    \includegraphics[width=\linewidth]{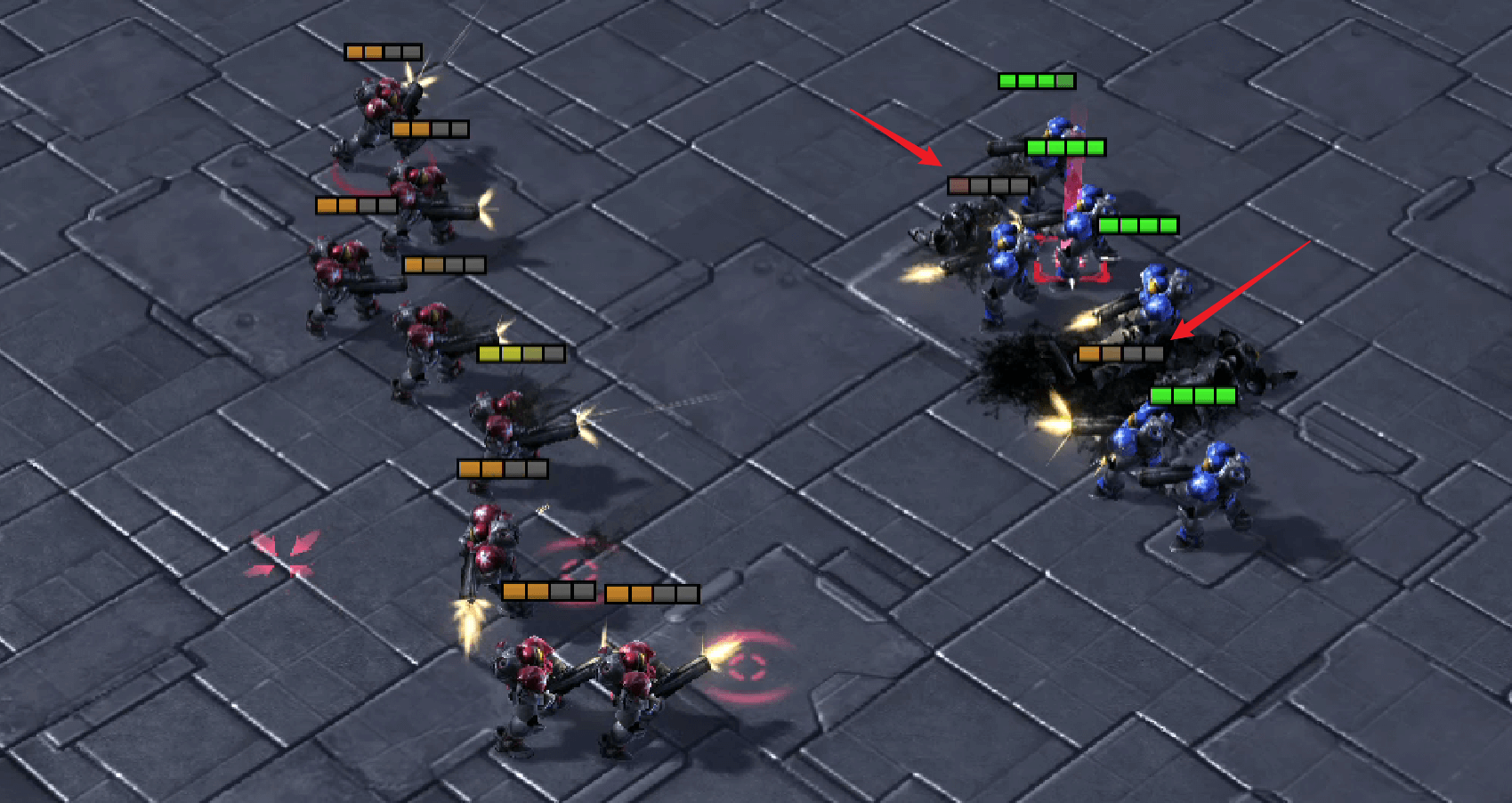}
    \label{appendix:8m9m_fire}
     \vspace{-8pt}
    \end{minipage}
    \vspace{-8pt}
    }
\vspace{-8pt}
    \caption{We denote agents trained on the Am\_vs\_Bm map and tested on the Cm\_vs\_Dm map as Train (A-B)-Test (C-D). For example, Train (5-6)-Test (8-9) means agents are tested on the 8m\_vs\_9m map while trained on the 5m\_vs\_6m map, corresponding to the settings of transfer learning. Train (5-6)-Test (5-6) means agents are trained and tested on the 5m\_vs\_6m map.
    \textbf{(a):} Agents' positions of Train (5-6)-Test (5-6).
    \textbf{(b):} Agents' positions of Train (5-6)-Test (8-9).
    As shown in the left half side of (a) and (b), ACE perfectly transfers the station position of 5m\_vs\_6m to 8m\_vs\_9m, that is, making the agents stand up and down in a line. 
    \textbf{(c):} Careful positioning and alternating fire of Train (5-6)-Test (8-9). ACE perfectly transfers the operation of careful positioning and alternating fire of 5m\_vs\_6m to 8m\_vs\_9m. Each ally has very little health but kills all enemies.
    \textbf{(d):} Focusing fire in Train (5-6)-Test (5-6).
    \textbf{(e):} Focusing fire in Train (5-6)-Test (8-9).
    \textbf{(f):} Focusing fire in Train (8-9)-Test (8-9).
    The optimal choice (d) in 5m\_vs\_6m is all allies focusing fire on one enemy while the optimal choice (f) in 8m\_vs\_9m is all allies focusing fire on two enemies, since 5 allies can kill an enemy instantly and having 8 agents attacking one agent (e) at the same time is wasteful.
}
\label{appendeix: transfer}
\vspace{-4ex}
\end{figure*}

\subsection{ACE-CTDE}
In this subsection we provide the detail of the implementation of $Q\left(o_i^t,a_i\right)$ in ACE-CTDE. For each agent $i$, we generate $Q\left(o_i^t,a_i\right)$ in the same way as that of $V\left(s_{a_{1:i}}^t\right)$, with the same unit encoder, action encoder and value encoder, except two differences. First, due to the partial observation, we only calculate the unit embedding of the units observed by agent $i$. Also, the edge feature of these units only includes the relation with the units observed by agent $i$. Second, due to the limited communication, only the action embedding $e_a\left(a_i\right)$ of agent $i$ is added on the corresponding unit embeddings, rather than all the preceding actions. 
Finally, the set of unit embeddings (only of the units observed by agent $i$) incorporated with only $e_a\left(a_i\right)$ is encoded by a value encoder to generate $Q\left(o_i^t,a_i\right)$, with the same structure as that for $V$.

\section{Additional Experiments}
\label{sec: add_exp}
\subsection{The behavior of transferred agents}
In this section, we carefully analyze the behavior of transferred agents. As shown in Figure~\ref{appendeix: transfer},  ACE perfectly transfers the station position, careful positioning, alternating fire, and focusing fire of 5m\_vs\_6m to 8m\_vs\_9m and then achieves 40\% win rate in 8m\_vs\_9m. However, the optimal choice of focusing fire in 8m\_vs\_9m is different from that in 5m\_vs\_6m, which is largely responsible for ACE not achieving a 100$\%$ win rate on the 8m\_vs\_9m map.
Inspired with ACE, in future work, we will design an algorithm to use only one policy to solve all maps in SMAC consisting of different unit types and numbers.

\begin{table}[h]
\centering
\resizebox{0.95\linewidth}{!}{
\begin{tabular}{cc}
\toprule
\multicolumn{1}{c|}{Parameter}        & Value                            \\ \midrule
\multicolumn{2}{c}{Exploration}                                       \\ \midrule
\multicolumn{1}{c|}{action\_selector}      & epsilon\_greedy                  \\
\multicolumn{1}{c|}{epsilon\_type}         & linear                           \\
\multicolumn{1}{c|}{epsilon\_start}        & 1                                \\
\multicolumn{1}{c|}{epsilon\_end}          & 0.05                             \\
\multicolumn{1}{c|}{\multirow{2}{*}{epsilon\_decay}}       
                        & 50k for academy\_3\_vs\_1\_with\_keeper \\
\multicolumn{1}{c|}{}   & 300k for academy\_counterattack\_hard                            \\ \midrule
\multicolumn{2}{c}{Sampler}                                           \\ \midrule
\multicolumn{1}{c|}{collector\_env\_num}   & 8                                \\
\multicolumn{1}{c|}{episode\_per\_collect}            & 32                               \\
\multicolumn{1}{c|}{replay\_buffer\_size}  & 300k                             \\ \midrule
\multicolumn{2}{c}{Training}                                          \\ \midrule
\multicolumn{1}{c|}{update\_per\_collect}  & 50                               \\
\multicolumn{1}{c|}{batch\_size}           & 320                              \\
\multicolumn{1}{c|}{weight\_decay}         & 1e-5                         \\
\multicolumn{1}{c|}{\multirow{2}{*}{learning\_rate}} 
       & 0.002 for academy\_3\_vs\_1\_with\_keeper \\
\multicolumn{1}{c|}{} 
        & 0.0009 for academy\_counterattack\_hard                         \\
\multicolumn{1}{c|}{target\_update\_theta} & 0.08                            \\
\multicolumn{1}{c|}{discount\_factor}      & 0.99                             \\
\multicolumn{1}{c|}{optimizer}             & rmsprop                          \\ \midrule
\multicolumn{2}{c}{Model}                                             \\ \midrule
\multicolumn{1}{c|}{hidden\_len}           & 128                              \\
\bottomrule
\end{tabular}}
\vspace{-0.2cm}
\caption{Hyperparameter Settings of ACE on GRF.}
\vspace{-3ex}
\label{appendix:grf_hyperparameter}
\end{table}

\end{document}